\documentclass[journal]{IEEEtran}
\usepackage{graphicx}
\usepackage{wrapfig}
\usepackage{amsmath}
\usepackage{subfig}
\usepackage{esvect}
\usepackage[utf8]{inputenc}
\usepackage{color}
\usepackage[square,sort,comma,numbers]{natbib}

\usepackage[numbers]{natbib}
\begin{document}

\title{Cascaded Region-based Densely Connected Network for Event Detection: A Seismic Application}

\author{Yue Wu$^1$,
        Youzuo Lin$^{1,*}$,
        Zheng Zhou$^{2}$,
        David Chas Bolton$^{3}$,
        Ji Liu$^{2}$, and Paul Johnson$^{1}$,
\thanks{\textbf{1}: the Earth and Environmental Sciences, Los Alamos National Laboratory, Los Alamos,
NM, 87544 USA.}
\thanks{\textbf{2}:  the Department of Computer Science, University of Rochester, Rochester, NY 14627.}
\thanks{\textbf{3}:  the Department of Geosciences, Penn State University, University Park, PA 16802.}
\thanks{\textbf{*}:  Correspondence to: Y. Lin, ylin@lanl.gov.}
\thanks{Manuscript received Nov, 2017.}}

\markboth{IEEE Transactions on Geoscience and Remote Sensing}%
{Shell \MakeLowercase{\textit{et al.}}: Bare Demo of IEEEtran.cls for IEEE Journals}

\maketitle

\begin{abstract}
Automatic event detection from time series signals has wide applications, such as abnormal event detection in video surveillance and event detection in geophysical data. Traditional detection methods detect events primarily by the use of similarity and correlation in data. Those methods can be inefficient and yield low accuracy. In recent years, because of the significantly increased computational power, machine learning techniques have revolutionized many science and engineering domains. In particular, the performance of object detection in 2D image data has been significantly improved due to the deep neural network. In this study, we apply a deep-learning-based method to the detection of events from time series seismic signals. However, a direct adaptation of the similar ideas from 2D object detection to our problem faces two challenges. The first challenge is that the duration of earthquake event varies significantly; The other is that the proposals generated are temporally correlated. To address these challenges, we propose a novel cascaded region-based convolutional neural network to capture earthquake events in different sizes, while incorporating contextual information to enrich features for each individual proposal. To achieve a better generalization performance, we use densely connected blocks as the backbone of our network. Because of the fact that some positive events are not correctly annotated, we further formulate the detection problem as a learning-from-noise problem. To verify the performance of our detection methods, we employ our methods to seismic data generated from a bi-axial ``earthquake machine'' located at Rock Mechanics Laboratory, and we acquire labels with the help of experts. Through our numerical tests, we show that our novel detection techniques yield high accuracy. Therefore, our novel deep-learning-based detection methods can potentially be powerful tools for locating events from time series data in various applications.
\end{abstract}


\IEEEpeerreviewmaketitle

\section{Introduction}

\IEEEPARstart{T}{ime} series data can be acquired through sensor-based monitoring. In the past few years, there have been increased interests to detect useful events out of various time series datasets for different applications. Among all these problems, seismic monitoring to detect the Earthquake has attracted many interests~\citep{Complexity-2016-Hsu, Earthquake-2015-Yoon}. In this study, we develop a novel event detection method and further employ our method to seismic time series datasets.

Machine learning methods have been successful in object detection to identify patterns. There have been many existing machine learning methods to detect events out of time series datasets in various applications such as epileptic seizure detection from EEG signals and change detection from remotely sensed imagery datasets. Depending on the availability of labeled datasets, all these event detection methods for time series date sets can be categorized into supervised~\citep{Event-2015-Oehmcke, MultiSensor-2015-Hassan, Batal-2012-Mining} and
unsupervised methods~\citep{Unsupervised-2017-Ahmad, Unsupervised-2016-Mur, Earthquake-2015-Yoon}. Our study belongs to the supervised category, since we acquire labels for training and evaluation with the help of experts. As for those supervised methods, they are all point-wised detection methods meaning they classify data point at each time stamp. Point-wised detection methods can be limited in their detection performance. In particular, those methods can neither accurately localize events nor obtain the number of events. In this study, inspired by the object detection in 2D imagery, we develop a novel event-wised detection method to capture each complete event. In other words, our detection methods capture the beginning and end coordinates to localize each event from the time series datasets.

Convolutional neural network~(CNN)  has achieved promising results in computer vision, image analysis, and many other domains due to the significantly improved computational power~(\cite{DenseNet, ResNet, VGG, AlexNet}). The State-of-the-Art CNN-based object detection models for 2D imagery mainly consists of two steps~\citep{FasterRCNN, FastRCNN, RCNN}: a step to generate the region proposals and a step to identify and localize the events within proposals. Specifically, segments of the input data that may include targeting patterns are first used to generate region proposals. A classifier is then employed on each proposal to detect targeting patterns, and a regressor is utilized to localize events within positive proposals. The original proposal generation method for CNN-based detection models is developed in two region-based CNN models, known as R-CNN~\citep{RCNN} and Fast R-CNN~\citep{FastRCNN}, where fixed methods are used to obtain proposals. Faster-RCNN~\citep{FasterRCNN} improves previous models by building region proposal networks~(RPN) on top of the final feature map of CNN backbone. Compared with \citet{FastRCNN} and \citet{RCNN}, the Faster-RCNN eliminates the additional time spent on proposal generation. To determine whether a proposal is positive or negative,  \citet{FasterRCNN} introduces \textit{anchor} to denote the region on the input data that a proposal covers. A proposal is considered positive if its corresponding anchor overlaps the ground-truth above a threshold.

In this study, we develop a novel deep neural network detection method for time series datasets. Similar to previous 2D detection models, our model also consists of two steps: proposal generation and event localization. However, a direct adaptation of 2D methods to generate the region proposals does not work well with our 1D seismic time series datasets because the duration of seismic events varies significantly. Therefore, we develop a novel region proposal method to address this issue. In particular, we develop a cascaded network that generates proposals at different by including more downsampling layers than regular networks do. Theoretically, events of small size can be captured at shallow layers. As the network becomes deeper, events of large size can be captured due to the increasing size of the receptive field. We add detection branches on feature maps at different depth.

Features are critical to the performance of our detection model. Since the classifier and regressor in the second step share the same feature vector obtained from CNN, enriching features for proposals will boost the detection rate and localization accuracy. Another novelty of our work is the incorporation of contextual information for each individual proposal. Although the importance of contextual information has been emphasized for imagery segmentation~\citep{Deeplabv3, PSPNet, GCN}, there is surprisingly few detection model taking into account contextual information on proposal level. As for our time series seismic signals, proposals are temporally correlated. Utilizing each proposal individually generates many false-positive detections. This is because these proposals may be part of some large events, and our detection method should be able to distinguish those small signal segments from large events. Considering this, we enrich features of each proposal by incorporating contextual information.

Because of the cascaded structure, the number of parameters in our model may significantly increase. To obtain a better generalization performance, we build our model based on densely connected network (DenseNet)~\citep{DenseNet}. The core idea of DenseNet is to reuse features learned from shallow layers, which enables to maintain a reasonable number of parameters even if the network becomes substantially deep. Another strategy we use to address overfitting is to share the parameters of the sibling detector and regressor. This is reasonable since we are interested in capturing specific patterns regardless of their sizes.

Another challenge of our event detection problem is that it is impractical for domain experts to annotate all events because the pattern of seismic events is not as obvious as the one in image objects. Those omitted events may bias the classifier for proposals. To alleviate the impact of mis-labeled positive events, we further formulate the proposal classification as a learning-from-noise problem. Inspired by \citet{LearningWithNoiseLabels}, we use a label-dependent loss function for the classifier.

We test our detection models on seismic time series  data and compare the experiment results obtained using the proposed cascaded contextual region-based CNN (CC-RCNN) and the traditional template matching method. We also conduct ablation experiments to verify the effect of the multi-scale architecture and the incorporation of contextual information. The experiment results demonstrate that our deep-learning-based model significantly outperform the template matching method, and the incorporation of contextual information for each individual proposal not only reduces false-positive detections, but also significantly increases the event localization accuracy. Also the utilization of label-dependent loss further boosts the performance of our detection models. To summarize, our contributions can be listed as follow:

\begin{itemize}
\item Extend region-based convolutional neural networks to time series scenarios;
\item Propose a cascaded structure to generate multi-scale proposals to efficiently capture events in varying lengths;
\item Incorporate contextual information for each proposal to further boost the detection accuracy;
\item Conduct experiments on seismic time series data and obtain promising results--achieving average precision~(AP)@[.50, .95] of $63.8\%$.
\end{itemize}

The rest of the paper is organized as follows. Section~\ref{sec:RelatedWork} briefly reviews related works on event detection and object detection.
Section~\ref{sec:Background} gives background knowledge about our method. The proposed method is elaborated in Section~\ref{sec:ProposedMethod}. Implementation details of the proposed methods and counterpart methods are described in Section~\ref{sec:ImplementationDetail}. Experiment results are provided in Section~\ref{sec:Experiment}. Section~\ref{sec:Conclusion} concludes this paper.

\section{Related Work}
\label{sec:RelatedWork}
Our study is related to both event detection for 1D time series datasets and object detection for 2D imagery datasets.
\subsubsection{Detection Methods for 1D Datasets}
There are many event detection methods in various applications. In seismology, STA/LTA is the most popular used detection method due to its simplicity~\cite{Automatic-1982-Allen, comparison-1998-Withers}. STA/LTA computes the ratio of short-term average energy and long-term average
energy on multiple receivers. If a seismic event is detected at a minimum of four stations, it is considered as an detection. However, STA/LTA fails to detect earthquakes or yields many false detections in challenging situations such as signal with low signal-to-noise ratio, or overlapping events. Compared to STA/LTA, Autocorrelation yields a much higher detection rate~\cite{autocorrelation-2008-Brown}. Autocorrelation is an exhaustive ``many-to-many'' detection method. It searches for similar waveforms when the desired signal waveform is unknown. The major disadvantage with autocorrelation
is its expensive computational cost. Its computational complexity scales quadratically with data duration. Template matching is a detection method method that yields a good balance between accuracy and computational complexity~\cite{Gibbons-2006-Gibbons,Nonvolcanic-2007-Shelly}. Template matching is a ``one-to-many'' detection method. It computes the correlation coefficient of a template waveform with the candidate waveform data. A detection is claimed when the correlation coefficient value is above the user-defined threshold value. Template matching has been proved to be efficient and successful for different seismic applications: microseismic monitoring in geothermal~\cite{Low-2013-Plenkers}, oil and gas reservoir~\cite{Improved-2010-Song}, nuclear monitoring~\cite{Perspectives-2014-Bobrov}, and tectonic tremor~\cite{Nonvolcanic-2007-Shelly}, etc.
\citet{Earthquake-2015-Yoon} recently developed an  unsupervised event detection approach called fingerprint and similarity thresholding~(FAST) method and apply it to detect earthquake out of seismic datasets. FAST  creates ``fingerprints'' of waveforms by extracting key discriminative features, then group similar fingerprints together within a database to facilitate the fast and scalable search for similar fingerprint pairs.

There are many other event detection methods developed in other application domains. \citet{Event-2015-Oehmcke} employed local outlier factor to detect events from marine time series data. To further improve results, dimensionality reduction methods are employed by the authors to the datasets. In the work of \citet{Batal-2012-Mining}, an event detection method was developed based on recent temporal patterns. The detection algorithm mines time-interval patterns backward in time, starting from patterns related to the most recent observation. The authors further applied their detection method to health care data of diabetic patients. \citet{Event-2007-McKenna} developed a binomial event discriminator~(BED) method. BED uses a failure model based on the binomial distribution to determine the probability of an event within a time segment. They applied the method to hydrological datasets to detect events.

\subsubsection{CNN-based Detection Methods for 2D data}
\citet{FasterRCNN} developed the faster RCNN method, of which a window is slid on the final feature map of the fourth stage of ResNet~\citep{ResNet} to generate proposals. The authors use nine different anchors with three various sizes (128, 256, 512) and three ratios of height/width (1:1, 1:2, 2:1) to determine regions that a proposal covers. To make anchors more accurate, \citet{UnifiedMultiScale} developed the multi-scale CNN methods, consisting of a proposal sub-network to generate multi-scale proposals at three stages of VGG~\citep{VGG} network. The authors then built detectors on top of each proposal branch.


\section{Background and Related Work}
\label{sec:Background}
\subsection{Template Matching}
Template matching (TM)~\citep{TemplateMatching} is widely used in the seismology community. It calculates the similarity of a template with successive windows from continuous waveform data. The commonly used similarity metric is normalized cross-correlation (CC),

\begin{equation}
\text{CC}(\mathbf{a}, \mathbf{b}) = \frac{\langle \mathbf{a}, \mathbf{b} \rangle}{\|\mathbf{a}\|_2 \|\mathbf{b}\|_2} = \frac{\sum_i \mathbf{a}_i \mathbf{b}_i}{\sqrt{\sum_i \mathbf{a}_i^2} \sqrt{\sum_i \mathbf{b}_i^2}},
\end{equation}

where $\mathbf{a}$, $\mathbf{b}$ are vectorized time series signals.
The detection threshold of $\tau$ is given as
\begin{equation}
\label{eq:tm_threshold}
 \tau = \mu \cdot \mathrm{~median~absolute~deviation~(MAD)},
\end{equation}
where $\mu$ is usually chosen as 9~\cite{TM_threshold_1,TM_threshold_2,TM_threshold_3}.
For a univariate set ${X_1, X_2, ..., X_n}$, MAD can be calculated as
\begin{equation}
\text{MAD}(X) = \text{median}(|X_i - \text{median}(X)|).
\end{equation}

\begin{figure}[!b]
	\centering
    \includegraphics[width=0.8\linewidth]{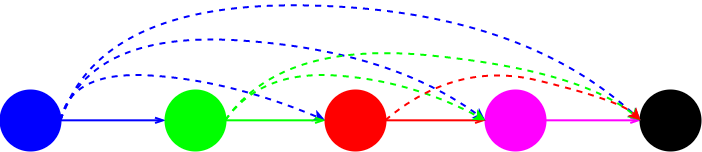}
    \caption{A densely connected block. Solid lines denote composite function $\mathcal{H}$, dashed lines denote concatenation.}
    \label{fig:densenet_block}
\end{figure}

\subsection{Densely Connected Network}

Densely connected network (DenseNet)~\citep{DenseNet} is an improved version of residual network~\citep{ResNet}. Both ResNet and DenseNet are discussed in this section.

\subsubsection{ResNet Block}
The major breakthrough of ResNet is the application of skip connections, which can be denoted as
\begin{equation}
x_{l + 1} = x_{l} + W'*(\sigma(B(W*(\sigma(B(x_{l}))
\label{eq:resnet_block}
\end{equation}
where $W$ and $W'$ are weight matrices, the operator of ``*'' denotes convolution, $B$ denotes batch normalization (BN)~\citep{BN}, $\sigma(x) = \text{max}(0, x)$ ~\citep{ReLU}. Eq.~\eqref{eq:resnet_block} forms a building block in ResNet. Skip connections are implemented by summing up the input of the block and the output of a set of convolution layers. The existence of skip connections weakens the importance of each individual path, so that the model behaves like an ensemble of small networks. Although ResNet immediately topped most of benchmarks, a few drawbacks still need to be addressed. 1) The number of parameters becomes extremely large with hundreds layers of convolutions. 2) The network may not benefit from ``going deep'' due to the gradient vanishing problem, as indicated in \citet{ResNetEnsemble}.

\subsubsection{DenseNet Block}
A DenseNet block is then formulated as
\begin{align}
x_{l + 1} & = \mathcal{H}([x_{0}, x_{1}, ..., x_{l}]) \\
\mathcal{H}(x) & = W*(\sigma(B(x))),
\end{align}
where $[x_{0}, x_{1}, ..., x_{l}]$ denotes the concatenation of all outputs of previous layers. A DenseNet block is illustrated in Fig.~\ref{fig:densenet_block}.

Thus, the output of a layer in one DenseNet block is \textit{densely connected} with outputs of all deeper layers in the same block by means of concatenation. Thus, it fully exploits the advantage of skip connections. Moreover, features from shallow layers are reused by deep layers, which reduces the number of parameters, and the gradient vanishing problem is further alleviated by the concatenation layers.

The feature dimension $d_{l + 1}$ of $x_{l + 1}$ is calculated as
\begin{equation}
d_{l} = d_{0} + k \cdot l,
\end{equation}
where k (a.k.a the \textit{growth rate}) is the number of filters used for each convolution layer.


\subsection{Atrous Convolution}

Atrous convolution convolves input nodes with a dilation rate $d$, denoting the stride for each convolved location on input nodes. The output node $y_i$ of an atrous convolution layer is calculated as
\begin{equation}
y_i = \sum_{k=1}^{K} x_{i + d\cdot k}\cdot w_k,
\end{equation}
where $k$ is the dimension, $x_i$ is the input node and $w \in R^K$ is the kernel.

The regular convolution can be seen as a special case of atrous convolution with $d = 1$. Atrous convolution was first proposed in \cite{Deeplabv1} to address the low-resolution problem caused by downsampling layers (pooling, convolution with stride, etc). Atrous convolution essentially involves distant information by covering larger regions of input signals while maintaining the same number of parameters.

\section{Proposed Methods}
\label{sec:ProposedMethod}
\subsection{Network Architecture}
\begin{table}[h]
\begin{center}
\begin{tabular}{ |c|c|c|c| }
\hline
Stage & Layers & Dim. & Anchor Size\\
\hline
Convolution & conv7, 24, /2 & L/2 $\times$ 24 & - \\
\hline
Pool& max-pool3, /2 & L/4 $\times$ 24 & - \\
\hline
$D_1$ & [conv3, 12] $\times$ 6 & L/4 $\times$ 96 & - \\
\hline
$T_1$ & avg-pool2, /2 & L/8 $\times$ 96 & - \\
\hline
$D_2$ & [conv3, 12] $\times$ 6 & L/8 $\times$ 168 & - \\
\hline
$T_2$ & avg-pool2, /2 & L/16 $\times$ 168 & - \\
\hline
$D_3$ & [conv3, 12] $\times$ 6 & L/16 $\times$ 240 & 128 \\
\hline
$T_3$ &
  \begin{tabular}{cc}
  conv1, 120\\
  avg-pool2, /2
  \end{tabular} & L/32 $\times$ 120 & - \\
\hline
$D_4$ & [conv3, 20] $\times$ 6 & L/32 $\times$ 240 & 256 \\
\hline
$T_4$ &
  \begin{tabular}{cc}
  conv1, 120\\
  avg-pool2, /2
  \end{tabular} & L/64 $\times$ 120 & - \\
\hline
$D_5$ & [conv3, 20] $\times$ 6 & L/64 $\times$ 240 & 512 \\
\hline
$T_5$ &
  \begin{tabular}{cc}
  conv1, 120\\
  avg-pool2, /2
  \end{tabular} & L/128 $\times$ 120 & - \\
\hline
$D_6$ & [conv3, 20] $\times$ 6 & L/128 $\times$ 240 & 1024 \\
\hline
$T_6$ &
  \begin{tabular}{cc}
  conv1, 120\\
  avg-pool2, /2
  \end{tabular} & L/256 $\times$ 120 & - \\
\hline
$D_7$ & [conv3, 20] $\times$ 6 & L/256 $\times$ 240 & 2048 \\
\hline
$T_7$ &
  \begin{tabular}{cc}
  conv1, 120\\
  avg-pool2, /2
  \end{tabular} & L/512 $\times$ 120 & - \\
\hline
$D_8$ & [conv3, 20] $\times$ 6 & L/512 $\times$ 240 & 4096 \\
\hline
$T_8$ &
  \begin{tabular}{cc}
  conv1, 120\\
  avg-pool2, /2
  \end{tabular} & L/1024 $\times$ 120 & - \\
\hline
$D_9$ & [conv3, 20] $\times$ 6 & L/1024 $\times$ 240 & 8192 \\
\hline
\end{tabular}
\caption{DenseNet architecture for event detection. Conv7, 64, /2 denotes using 64 $1 \times 7$ convolution kernels with stride 2. The same routine applies to max-pool and avg-pool. L denotes the length of input waveform. ``-'' means the output of that stage is not used to make predictions.}
\label{table:densenet_architect}
\end{center}
\end{table}

As illustrated in Table~\ref{table:densenet_architect}, our network is inspired by DenseNet. All convolution kernels in our network are 1D because of the input of 1D time series data. The brackets denote DenseNet blocks illustrated in Fig.~\ref{fig:densenet_block}. $D_i$, $T_i$ denote DenseNet blocks and transition blocks, respectively. All transition blocks have an average pooling layer to downsample the signal by 2, while $T_3 - T_9$ have an extra $1 \times 1$ convolution layer to reduce the feature dimension by half. As previously discussed, our model is designed for capturing events with significantly various durations, so we use the output of $D_3 - D_9$, having strides 16-1024 on the input signals, as proposals, then detection branches (the classification and regression layer) are built on the top of these multi-scale proposals. Since small events greatly outnumber large events, we share the detection branches for all scales to make our model robust. To achieve this, We set growth rate $k = 12$ for ${D_1, D_2}$, $k = 20$ for $D_3 - D_9$, so that all proposals have the same feature dimension 240.

\subsection{Anchors}
Anchor is the effective region of the input signals that a proposal is responsible for. In most cases, it is used to decide the label for that proposal. In our time series data, an anchor indicates two coordinates representing the beginning and the end of each proposal. We assign the anchor size 128 to proposals in $D_3$, and it doubles for the next scale. Proposals in $D_9$ have the largest anchor size 8192. These settings are determined by the length distribution of events in our data. It is worthwhile to mention that the amount of shifts between adjacent proposals is determined by the stride of that stage on the input signals. For example, the shift between two adjacent anchors of $D_3$ is 16 timestamps.


\begin{figure}[t]
\centering
\subfloat[]{\includegraphics[width=0.7\linewidth, height=3cm]{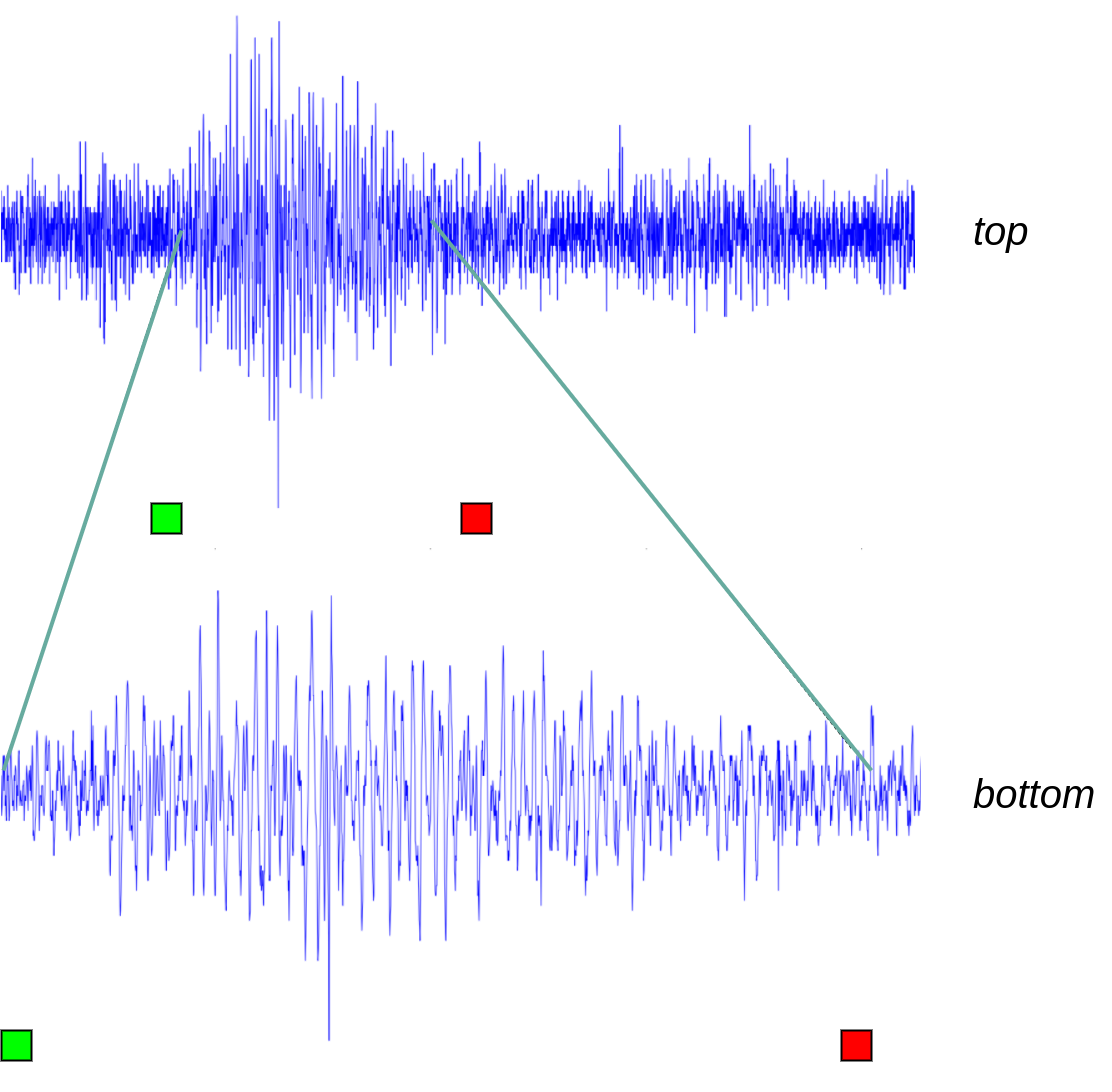}%
\label{fig:observe2}}
\hfil
\subfloat[]{\includegraphics[width=0.7\linewidth, height=3cm]{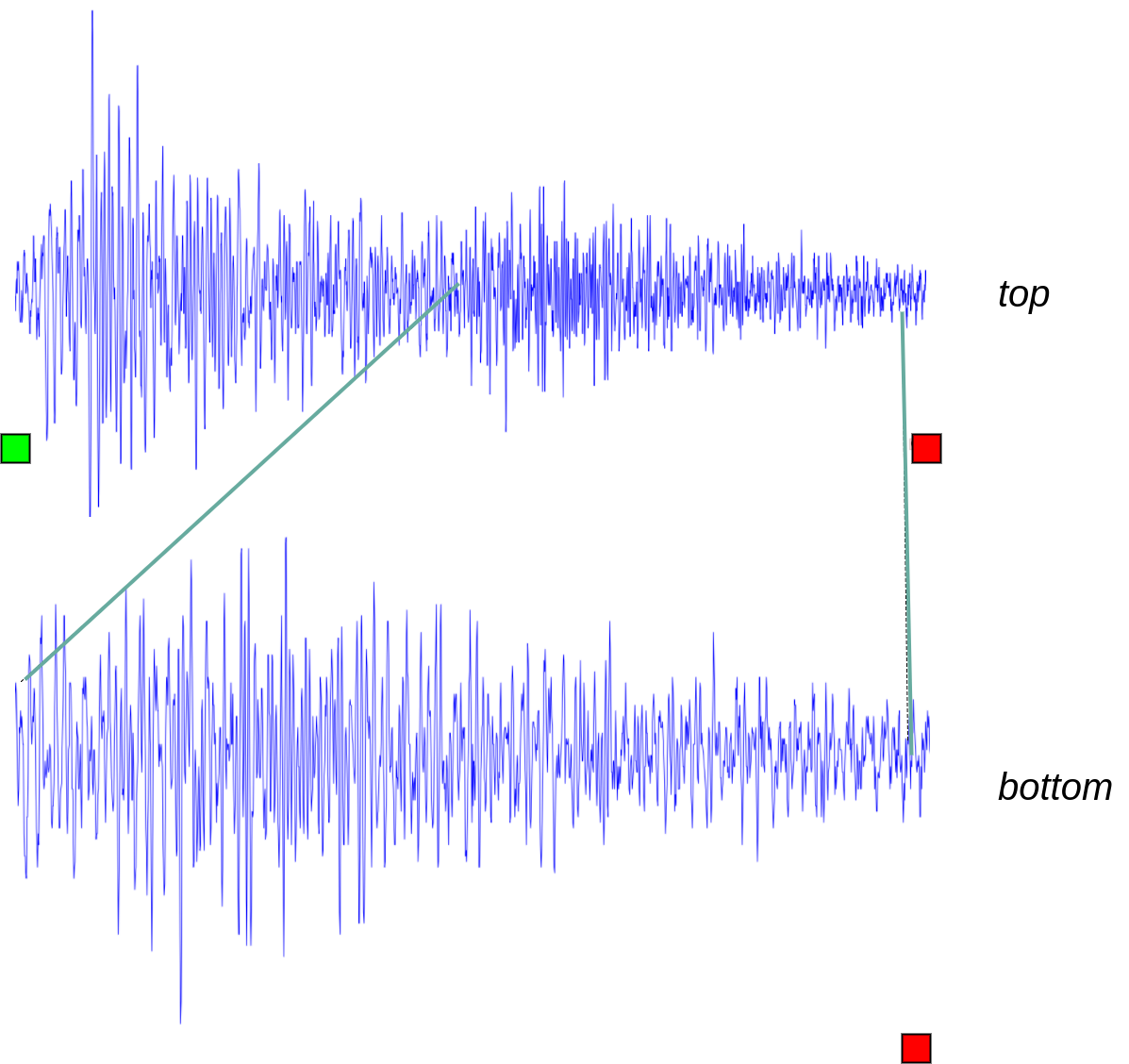}%
\label{fig:observe1}}
\caption{(a) \textit{top}: A perfect individual event in time series data. (a) \textit{bottom}: A zoomed-in truncation of
(a) \textit{top}.
(b) \textit{top}: A segment of time series signals, and the whole segment is
labeled as an event. (b) \textit{bottom}: A truncation of (b) \textit{top}.
The green and red boxes indicate the beginning and end of an event, respectively.}
\label{fig:observe}
\end{figure}

\subsection{Proposals with Contextual Information}
\label{sec:ContextualInfo}
Features are critical to detection. In time series data, it is important to take into consideration of temporal correlations among neighboring proposals. Considering features from each individual proposal only will result in many false detections.

Figure~\ref{fig:observe2} illustrates a perfect individual event in time series data. The signal amplitude keeps at the consistent level before a major event comes. As the event vanishes, the signal amplitude decreases to the previous level. However, it can be possible that the signal amplitude does not decrease monotonically or the major event may last longer than usual. Both cases will lead to the scenario when truncations from a major event are mis-detected as several small events. Figure~\ref{fig:observe1} illustrates an example of a false detection. The whole segment of signals in Fig.~\ref{fig:observe1} (denoted as ``top'') is a single event. However, if we only focus on a truncation of that, i.e., the ``bottom'' part as shown in Fig.~\ref{fig:observe1}, we may mistakenly consider this truncation as an individual event. Therefore, in order to detect each event as a whole, it is necessary to check preceding and succeeding patterns for each proposal.


\begin{figure}[t]
    \centering
    \includegraphics[width=\linewidth]{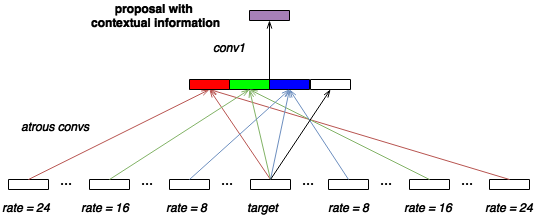}
    \caption{The atrous convolution block. We aim at enriching features of individual proposals by convolving with proceeding and succeeding proposals. In this figure, the target proposal is at the center. Atrous convolutions with different strides are applied to capture contextual information from nearby to further proposals.}
    \label{fig:aconv_color}
\end{figure}

We build atrous convolution blocks on seven proposal layers, $D_3$-$D_9$. The atrous convolution block is illustrated in Fig.~\ref{fig:aconv_color}. The dilation rate in atrous convolution indicates the number of skipped proposals at each convolved location. We set dilation rates to be 4, 8, and 12 for proposals in all scales. These dilation rates are inspired by the amount of shifts of adjacent proposals. Anchors of adjacent proposals shift only a little, hence the features of adjacent proposals tend to be similar. In contrast, atrous convolution is capable of incorporating contextual information. To the other extreme, the shifts of anchors should not be too large since the information from far away will be irrelevant to the target proposal. With dilation rates of 4, 8, 12, the shifts between the target proposal and contextual proposals are 0.5, 1, 1.5 of the anchor size. The blocks in red, green, and blue shown in Fig.~\ref{fig:aconv_color} are the outputs of atrous convolutions with 4, 8, and 12 as dilation rates, followed by batch normalization and activation layers. All convolutions are $1\times 3$, with 240 kernels. We generate new proposals with contextual information by concatenating outputs using three dilation rates and the target proposal. The new proposal includes four times as many features as the target proposal. In order to keep the number of features unchanged, we further employ a $1 \times 1$ convolution layer. To summarize, we employed atrous convolution with 3 dilation rates on proposals layers so that the features of each individual proposal are enriched by proceeding and succeeding proposals, while maintaining its own features.

\subsection{Sibling Branches for Detection and Localization}
\begin{figure}[h]
    \centering
    \includegraphics[width=0.9\linewidth]{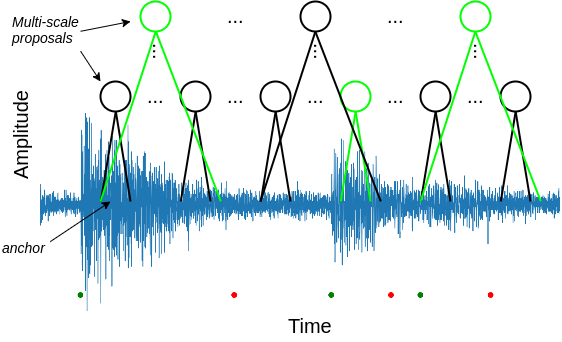}
    \caption{An illustration of multi-scale detections. There are three proposals (green nodes) should be considered positive since they have the intersection over union~(IoU) with the ground-truth above the threshold. Ground-truth are indicated at the bottom with green and red dots, which denote beginnings and ends, respectively.}
    \label{fig:detection_branch}
\end{figure}
We add a classification branch and a regression branch on each proposal. The classification branch is first used to detect whether a proposal includes an event or not. For each positive proposal, we further apply a regressor to localize the event within. We use a joint loss function to optimize classification and regression branches simultaneously.

We assign a positive label to a proposal if its anchor has the ratio of intersection over union (IoU) above 0.5 with at least one ground-truth event. Proposals are assigned a negative label if the highest IoU of their anchors with the ground-truth is below 0.3. Neutral proposals (IoU $\in$ [0.3, 0.5]) do not contribute to the loss. To localize the event within a proposal, two offsets: $d_x$ and $d_w$ are captured to transform the anchor to real coordinates by
\begin{equation}\label{eq:to_real_coordinates_1}
    G^{*}_x = P_w d_x + P_x,
\end{equation}
\begin{equation}\label{eq:to_real_coordinates_2}
    G^{*}_w = P_w exp(d_w),
\end{equation}
where $P_x$, $P_w$ are the center and length of an anchor, $G^{*}_x$, $G^{*}_w$ are the center and length of the prediction. Figure~\ref{fig:detection_branch} gives an illustration of this process, where three nodes are positive so they also have a localization loss.

Another challenge in our time series data is that not all events in the training set are annotated, which is caused by the fact that some patterns are difficult for our annotators to decide.  Due to this problem, negative labels are noisy in our task. To address this issue, we employ label-dependent cost function for the classifier.

Label-dependent cost function was initially proposed in a couple of work~\citep{Weighted_LogLoss, BiasedSVM}, which is known as weighted logistic regression and biased support vector machine, respectively. The core idea of ``label-dependent'' is to apply separate loss functions for positive and negative sets
\begin{equation}
J(g(x)) = \frac{1}{|X|}\left(\alpha \sum_{x \in X_+} l(g(x)) + \beta \sum_{x \in X_-} l(g(x))\right),
\label{eq:label_dependent_loss}
\end{equation}
where $l$ can be any 0-1 loss functions, $X_{+}$, $X_{-}$ denote the observed positive and negative sets, $\alpha$ and $\beta$ are two hyper-parameters, and $g$ is a linear score function.

To obtain the optimal weight parameters $\alpha^{*}$ and $\beta^{*}$, \cite{LearningWithNoiseLabels} set $\rho_{+1}=P(\tilde{Y} = -1|Y = 1)$ and $\rho_{-1} = P(\tilde{Y} = 1|Y = -1)$ to calculate
\begin{equation}
\alpha^{*} = \frac{1 - \rho_{+1} + \rho_{-1}}{2},
\label{eq:alpha_noise_relation}
\end{equation}
 and $\beta^{*} = 1 - \alpha^{*}$.
It can be shown that by employing the optimal parameters of $\alpha^{*}$ and $\beta^{*}$, the resulting classifier can make predictions of $sign(g(x) - 1/2)$ with noisy data~\citep{LearningWithNoiseLabels}. We use the similar parameter estimation approach to our datasets by setting $\rho_{+1} = 0$, since the noise only exists in negative samples.

\subsubsection{Loss Function}
We develop a joint loss function $L$ including a classification cost function $L_{cls}$ and a regression cost function $L_{regr}$
\begin{multline}\label{eq:loss_total}
L(d_{cls}, d_{x}, d_{w}, t_{cls}, t_{x}, t_{w}) = L_{cls}(d_{cls}, t_{cls}) + \\ \lambda 1\{t_{cls} = 1\} \sum_{u \in \{x, w\}}L_{regr}(d_{u}, t_{u}),
\end{multline}
where $1\{t_{cls} = 1\}$ is the indicator function indicating only positive proposals contribute to the regression loss, and $d_{cls}^{(i)}$, $d_{x}^{(i)}$, and $d_{w}^{(i)}$ are the predictions of the $i^{th}$ proposal's class score, center and length offsets, respectively, and $t_{cls}^{(i)}$, $t_{x}^{(i)}$, $t_{w}^{(i)}$ are the corresponding ground-truth of the $i^{\mathrm{th}}$ proposal's class score, center and length offsets, respectively, and $\lambda$ is the regularization parameter.

The classification cost function $L_{cls}$ is defined as a label-dependent logistic loss
\begin{multline}\label{eq:loss_cls_ld}
    L_{cls}(d_{cls}, t_{cls}) = \alpha 1\{t_{cls} = 1\}\log(1 + e^{-d_{cls}}) +\\ (1-\alpha) 1\{t_{cls} = -1\}\log(1 + e^{d_{cls}}),
\end{multline}
where $\alpha$ is the hyper-parameter.

The regression cost function $L_{regr}$ is defined as the smoothed $L_1$ loss as proposed in \cite{FastRCNN}:
\begin{gather}
L_{regr}(d_{u}, t_{u}) = \text{smooth}_{L_1}(t_{u} - d_{u}),
\end{gather}
where $\text{smooth}_{L_1}(x) =
  \begin{cases}
    0.5x^2 & \text{if $|x| < 1$} \\
    |x| - 0.5 & \text{otherwise}
  \end{cases}.$

According to Eqs.~\eqref{eq:to_real_coordinates_1} and \eqref{eq:to_real_coordinates_2}, $t_x$ and $t_w$ can be obtained by
\begin{align}
    t_x & = (G_x - P_x) / P_w,\\
    t_w & = ln(G_w/P_w).
\end{align}

\subsubsection{Share Weights for Robustness}
To capture events with dramatically varying durations, we make multi-scale predictions on output layers with different sizes of receptive fields. However, events with different lengths are not equally distributed. In other words, small events greatly outnumber large events. Our model should capture patterns from all events regardless of their durations or magnitude, hence we share the weights of contextual atrous convolution layers, sibling classification and regression branches built on top of $P_4$-$P_8$. Weight sharing makes our model robust and help with the optimization because predictions in all scales equally contribute to the loss function.



\section{Implementation Details}
\label{sec:ImplementationDetail}
\subsection{Template Matching}
We use events in the training set as templates. For each template, CC is calculated at each sliding location of the time series data. We set the detection threshold $\mu = 8$ in Eq.~\eqref{eq:tm_threshold}, which is determined by the validation set. For multi-detections of a single event, the detection with the highest CC is kept and all other detections are discarded. The beginning and the end of each detection are determined by those of the template.
\subsection{Proposed Model}
\subsubsection{Optimization}
The proposed model has approximately 3 million parameters. For each mini-batch iteration, we feed a 24,576-timestamp time series segment with 0.5 overlapping rate so that if the end point of an event lies outside of the segment, that event will be roughly at the center of the next segment. Table~\ref{table:selected_proposal} shows how we select proposals from each detection branch.

\begin{table}[h]
\centering
\begin{tabular}{ |c|c|c|c|c|c|c|c| }
\hline
stage & $C_3$ & $C_4$ & $C_5$ & $C_6$ & $C_7$ & $C_8$ & $C_9$ \\
\hline
\# of proposals & 64 & 64 & 64 & 64 & 32 & 32 & 16 \\
\hline
\end{tabular}
\caption{Number of proposals selected from each scale.}
\label{table:selected_proposal}
\end{table}

To further simplify the optimization, we also make sure the ratio of positive and negative proposals is 1:1. If positive or negative proposals are insufficient, we use neutral ones as negative proposals. Adam optimizer \citep{Adam} is applied with the initial learning rate of 5e-4. The learning rate is multiplied by 0.1 for every ten epochs. Each mini-batch data is subtracted by mean and divided by standard deviation before feeding into the network. The implementation is built on TensorFlow~\citep{tensorflow}.

\subsubsection{Inference}
Same as the training process, we feed 24,576 timestamp segments with the overlapping rate of 0.5 each time. Predictions are first generated on all proposes, and then apply non-maximum suppression (NMS) to reduce multi-detections for a single event. Since events in time series data are rarely overlapped, we set the IoU threshold of NMS to be 0.05.

\section{Experiment}
\label{sec:Experiment}
\subsection{Data}

We use acoustic time series data acquired at the Rock and Sediment Mechanics Laboratory of Penn State University. The dataset is a time-amplitude representation of acoustic emissions generated by a double-direct shearing apparatus ~\citep{Laboratory-2016-Leeman, Effect-1998-Karner}. There are 3,357,566 timestamps in total, spanning approximately 0.9 seconds. 1000 seismic events are manually picked by experts. We use 800 events for training, 100 events for validating, and 100 events for testing.

\begin{figure}[t]
    \centering
    \includegraphics[width=0.8\linewidth]{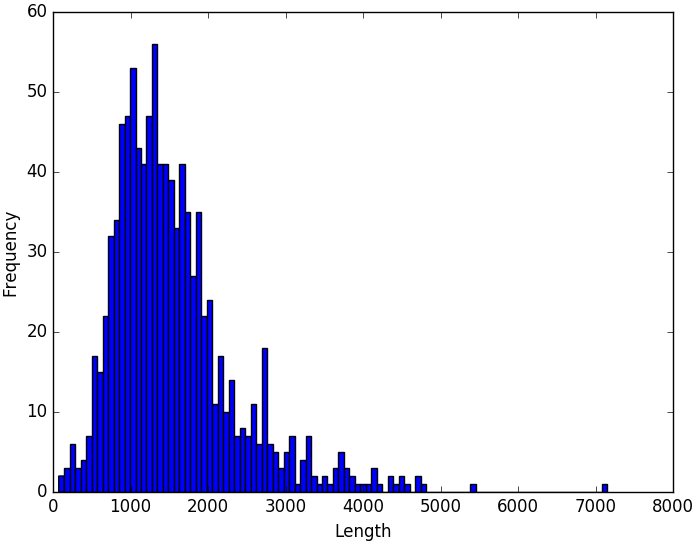}
    \caption{The length distribution of all events.}
    \label{fig:length_distribution}
\end{figure}

We calculate the length of all events. The length distribution is shown in Fig. \ref{fig:length_distribution}. The largest event spans more than 7,000 timestamps while the smallest event spans few hundreds. The mean and median lengths of events are both about 1,500 timestamps.

\subsection{Metric}
We use average precision~(AP) to evaluate our models. AP first calculates the precision-recall curve, then averages maximum precisions for each unique recall. For AP@.5, a detection is considered as true positive if it has IoU above 0.5 with a ground-truth event. If there are multiple detections for one event, only one detection is considered true positive, others are considered false positive. In this study, we use AP@[.5, .95], which is used in MS COCO object detection dataset~\citep{MSCOCO}.
To obtain AP@[.5, .95], we calculate 10 APs using IoUs from 0.5 to 0.95 with stride 0.05, then take the average of 10 APs. The IoU for two 1D segments $\overline{A_0A_1}$, $\overline{B_0B_1}$ is calculated as:
\begin{gather*}
x_a = \max(A_{0}, B_{0}),\\
y_a = \min(A_{1}, B_{1}),\\
x_b = \min(A_{0}, B_{0}),\\
y_b = \max(A_{1}, B_{1}),\\
IoU(\overline{A_0A_1}, \overline{B_0B_1}) = \frac{\max(y_a - x_a, 0)}{y_b - x_b}.
\end{gather*}

\subsection{Test: Overall Performance}

\begin{table}[h]
\centering
\begin{tabular}{ c|c|c }
& TM & CC-RCNN \\
\hline
AP@.50 & 22.0\% & 95.7\% \\
AP@.55 & 12.4\% & 94.6\% \\
AP@.60 & 7.4\% & 91.5\% \\
AP@.65 & 4.6\% & 87.5\% \\
AP@.70 & 3.8\% & 80.0\% \\
AP@.75 & 2.4\% & 72.5\% \\
AP@.80 & 1.1\% & 61.5\% \\
AP@.85 & 0.5\% & 39.2\% \\
AP@.90 & 0.3\% & 15.1\% \\
AP@.95 & 0.1\% & 0.7\% \\
\hline
AP@[.50, .95] & 5.5\% & \textbf{63.8\%} \\
\end{tabular}
\caption{Accuracy results obtained using template matching and our CC-RCNN model.
We notice that our CC-RCNN model consistently yields better detection accuracy than the template matching method. }
\label{table:results_tm_crcnn_ccrcnn}
\end{table}

\begin{figure}[!b]
\centering
\subfloat[]{\includegraphics[width=0.9\linewidth]{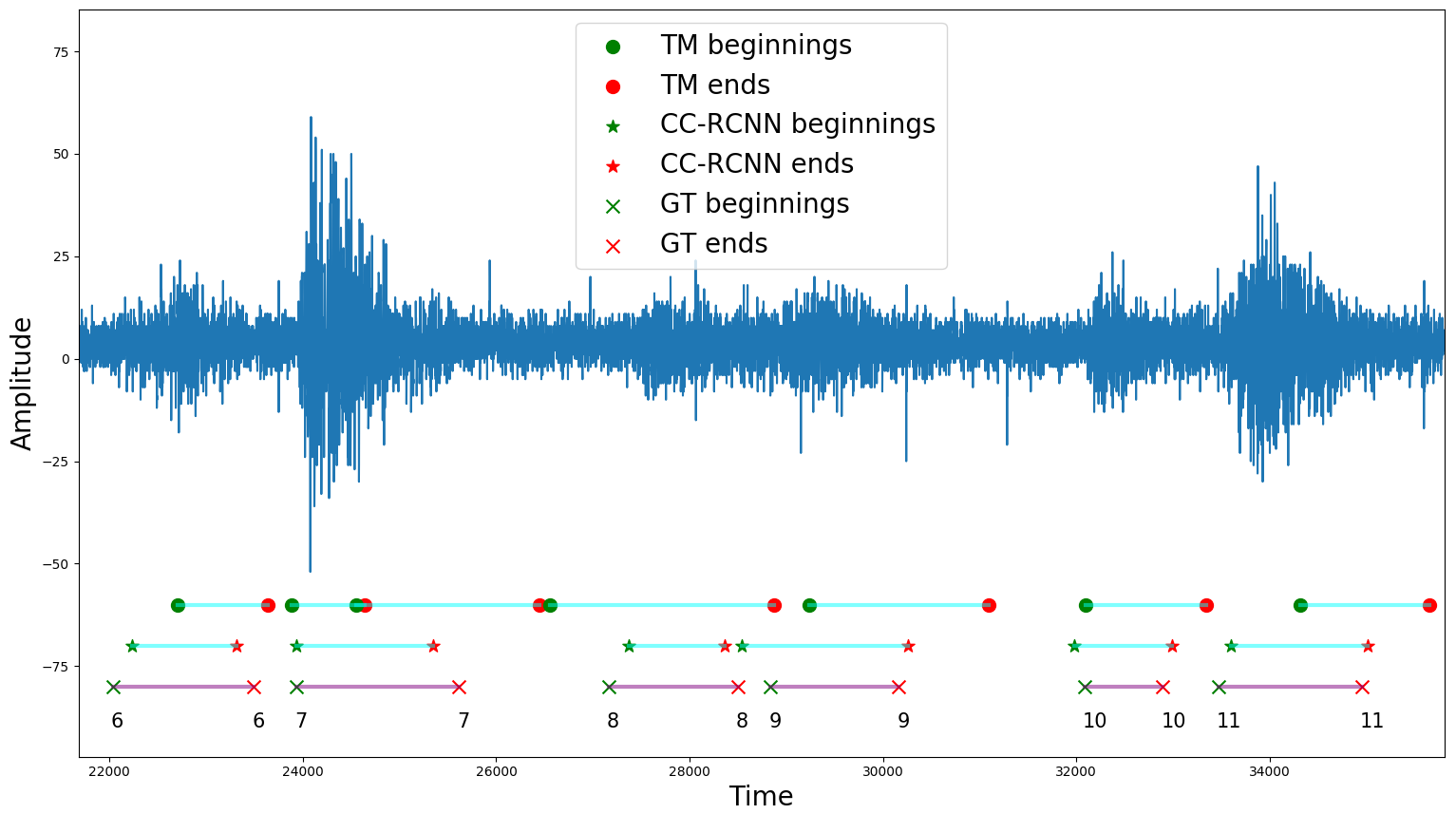}%
\label{fig:tm_ccrcnn_a}}
\hfil
\subfloat[]{\includegraphics[width=0.9\linewidth]{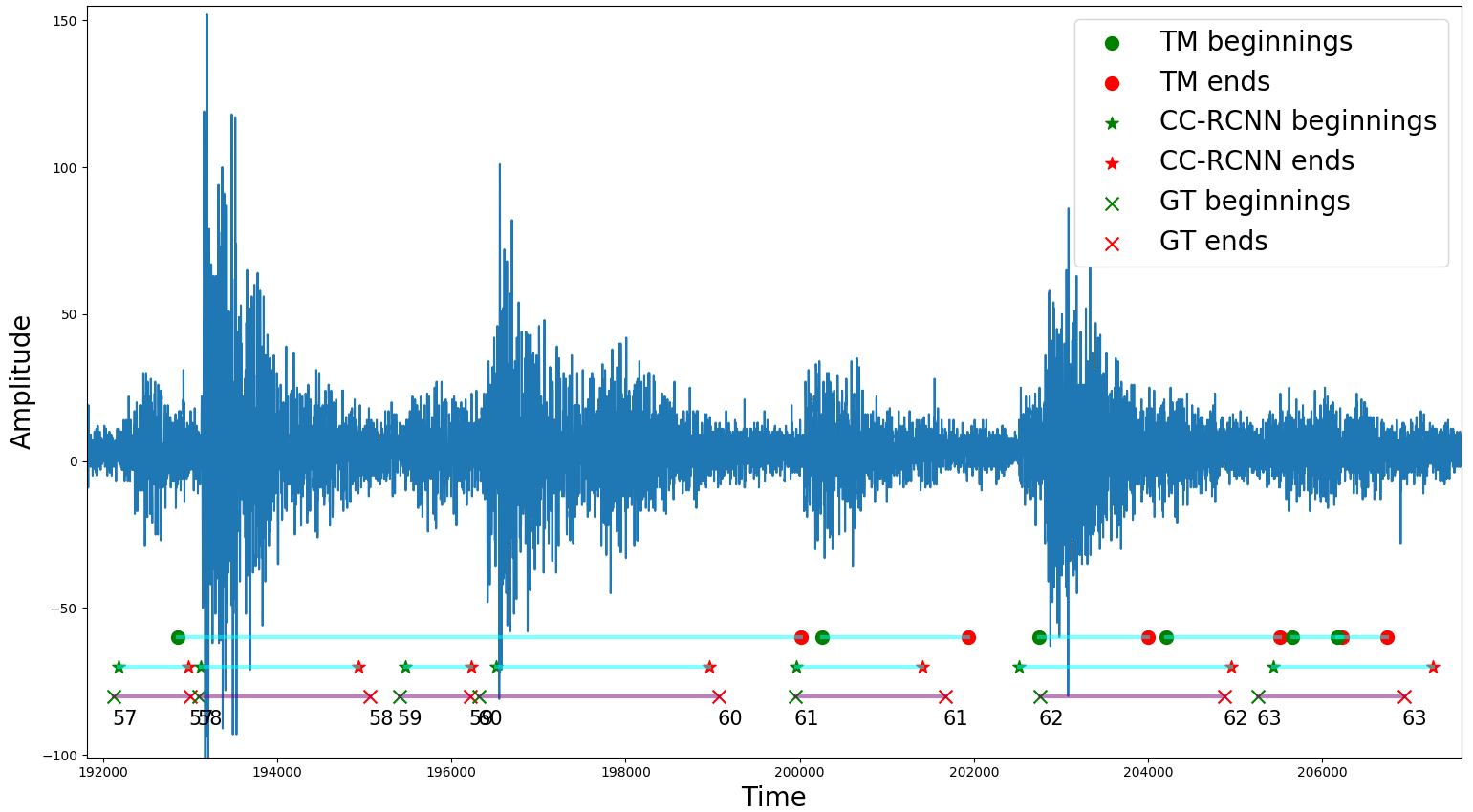}%
\label{fig:tm_ccrcnn_b}}
\caption{We show two different examples in (a) and (b).
To capture events shown in (b) is more challenging than those in (a) since some of events in (b) vary dramatically in
lengths (such as \#57, \#58) and yields irregular patterns (such as \#60). We compare the results obtained using our method (denoted by ``$\star$'')
to those obtained by using template matching (``$\bullet$''). The ground-truth (``$\times$'') is also provided.
Our detection methods yield much higher accuracy than template matching method.}
\label{fig:tm_ccrcnn}
\end{figure}

We provide detection results in Table~\ref{table:results_tm_crcnn_ccrcnn} obtained using the template matching (baseline), and our deep-learning-based CC-RCNN.
Table~\ref{table:results_tm_crcnn_ccrcnn} shows that our CC-RCNN model consistently yields better detection accuracy than the template matching method.
The average AP value of our method is $63.8\%$ compared to the value of $5.5\%$ obtained by the TM method.

To visualize the detection results using our method and the TM method, we provide the example detections of these two models in Fig.~\ref{fig:tm_ccrcnn}.
The results obtained using our method are denoted by ``$\star$'', and those obtained by template matching are denoted in ``$\bullet$''.
The ground-truth (``$\times$'') is also provided.  Although TM has the ability to detect some events, it has poor localization performance. For instance,
TM only captures the second half part of event \#6 in Fig.~\ref{fig:tm_ccrcnn_a}. Also, TM cannot capture event \#7 as a whole since each template
works separately. The reason for the poor localization is that the lengths of TM detections are simply determined by those of templates, and
most events have lengths that no templates can match. On the other hand, the proposed CNN-based method is capable of accurately detecting, as well as localizing multi-scale events due to the cascaded design.
To capture events shown in Fig.~\ref{fig:tm_ccrcnn_b} is more challenging than those in
Fig.~\ref{fig:tm_ccrcnn_a} since some of events in Fig.~\ref{fig:tm_ccrcnn_b} are varying dramatically in lengths (such as \#57, \#58) and yields irregular
patterns (such as \#60). TM detect \#57 - \#60 as a whole event, while CC-RCNN is able to detect and localize
each individual event accurately.

Based on results shown in Table.~\ref{table:results_tm_crcnn_ccrcnn} and Fig.~\ref{fig:tm_ccrcnn}, our detection methods yield much higher
accuracy than template matching method.

\subsection{Test: Hyper-Parameters}

\begin{table}[h]
\centering
\begin{tabular}{ c|c|c|c|c }
$\lambda$ & 0.1 & 1 & 10 & 100\\
\hline
AP@[.50, .95] & 48.2\% & 60.4\% & \textbf{61.5}\% & 56.0\% \\
\end{tabular}
\caption{Accuracy w.r.t different values of $\lambda$ in Eq.~\eqref{eq:loss_total}.
We test $\lambda \in \{0.1, 1, 10, 100\}$. The accuracy peaks when $\lambda=10$.}
\label{table:results_lambda}
\end{table}

\begin{table}[!b]
\centering
\begin{tabular}{ c|c|c|c|c|c }
& $\alpha=0.45$ & $\alpha=0.5$ & $\alpha=0.55$ & $\alpha=0.6$ & $\alpha=0.7$\\
\hline
AP@.50 & 97.3\% & 97.3\% & 95.7\% & 96.2\% & 93.0\%\\
AP@.55 & 95.0\% & 96.0\% & 94.6\% & 95.3\% & 92.0\%\\
AP@.60 & 89.0\% & 93.8\% & 91.5\% & 92.0\% & 87.2\%\\
AP@.65 & 83.7\% & 84.5\% & 87.5\% & 84.4\% & 82.2\%\\
AP@.70 & 78.4\% & 81.0\% & 80.0\% & 77.2\% & 78.8\%\\
AP@.75 & 71.5\% & 67.3\% & 72.5\% & 73.3\% & 68.6\%\\
AP@.80 & 51.6\% & 51.9\% & 61.5\% & 56.7\% & 54.8\%\\
AP@.85 & 28.3\% & 30.4\% & 39.2\% & 29.2\% & 35.5\%\\
AP@.90 & 11.1\% & 11.2\% & 15.1\% & 11.7\% & 13.7\%\\
AP@.95 & 0.9\% & 1.5\% & 0.7\% & 1.3\% & 1.0\%\\
\hline
AP@[.50, .95] & 60.7\% & 61.5\% & \textbf{63.8}\% & 60.7\% & 60.3\%\\
\end{tabular}
\caption{Accuracy w.r.t different values of $\alpha$ in Eq.~\eqref{eq:loss_cls_ld}.
We test $\alpha \in \{0.45, 0.5, 0.55, 0.6, 0.7\}$. The accuracy peaks at $\alpha = 0.55$, which has the corresponding noise level of 0.1.
The accuracy then decreases as $\alpha$ becomes larger. The best average performance of our CC-RCNN model is achieved when $\alpha=0.55$.}
\label{table:results_ld}
\end{table}

Hyper-parameters play an important role in our model to achieve high performance.
Specifically, the selection of $\lambda$ value in Eq.~\eqref{eq:loss_total} and  $\alpha$ value in Eq.~\eqref{eq:alpha_noise_relation} are
critical to the detection accuracy using our label-dependent loss function.

We provide results in Table~\ref{table:results_lambda} to illustrate the performance of our algorithm using different $\lambda$ values in Eq.~\eqref{eq:loss_total}.
Similar to \citet{FasterRCNN}, different values of $\lambda \in \{0.1, 1, 10, 100\}$~are tested. We observe that the performance of our model
can be impacted notably by using different $\lambda$ values. The best performance is achieved with $\lambda=10$, which is therefore used for
the all the tests implemented within the work.

To demonstrate the effectiveness of our selected $\alpha$ value, we further provide detection results using our label-dependent loss function
in Eq.~\eqref{eq:loss_cls_ld}.
The performance on different noise parameters, $\alpha \in \{0.45, 0.5, 0.55, 0.6, 0.7\}$,  is reported in Table.~\ref{table:results_ld}.
The best average performance of our CC-RCNN model is achieved when $\alpha=0.55$ with the corresponding noise level of 0.1.
This is reasonable since the noise only exists in negative samples, and noise level is low. The accuracy decreases when either the noise level
becomes larger ($\alpha=0.55, 0.6, 0.7$) or assuming that the noise exists in positive samples ($\alpha=0.45$).

Through these sets of tests on hyper-parameter, we obtain the best combination to use for our dataset i.e.,
$\lambda = 10$ and $\alpha = 0.55$.

\begin{figure}[!b]
\centering
\includegraphics[width=0.9\linewidth]{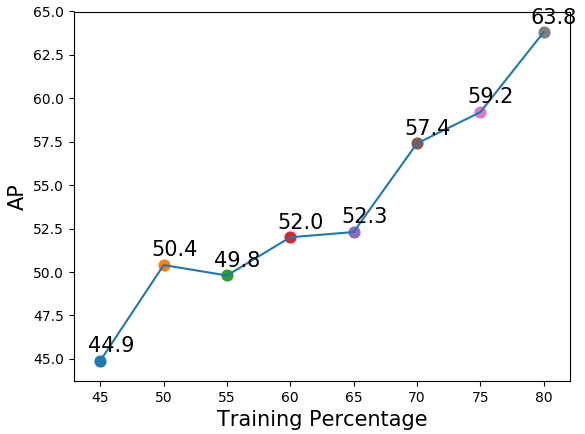}
\caption{The APs achieved by using different training data sizes. The AP peaks when using 800 samples for training. Even with 450 samples, which is less than a half, our model still has a good performance.}
\label{fig:trainsize}
\end{figure}

\begin{figure}[t]
\centering
\subfloat[]{\includegraphics[width=\linewidth]{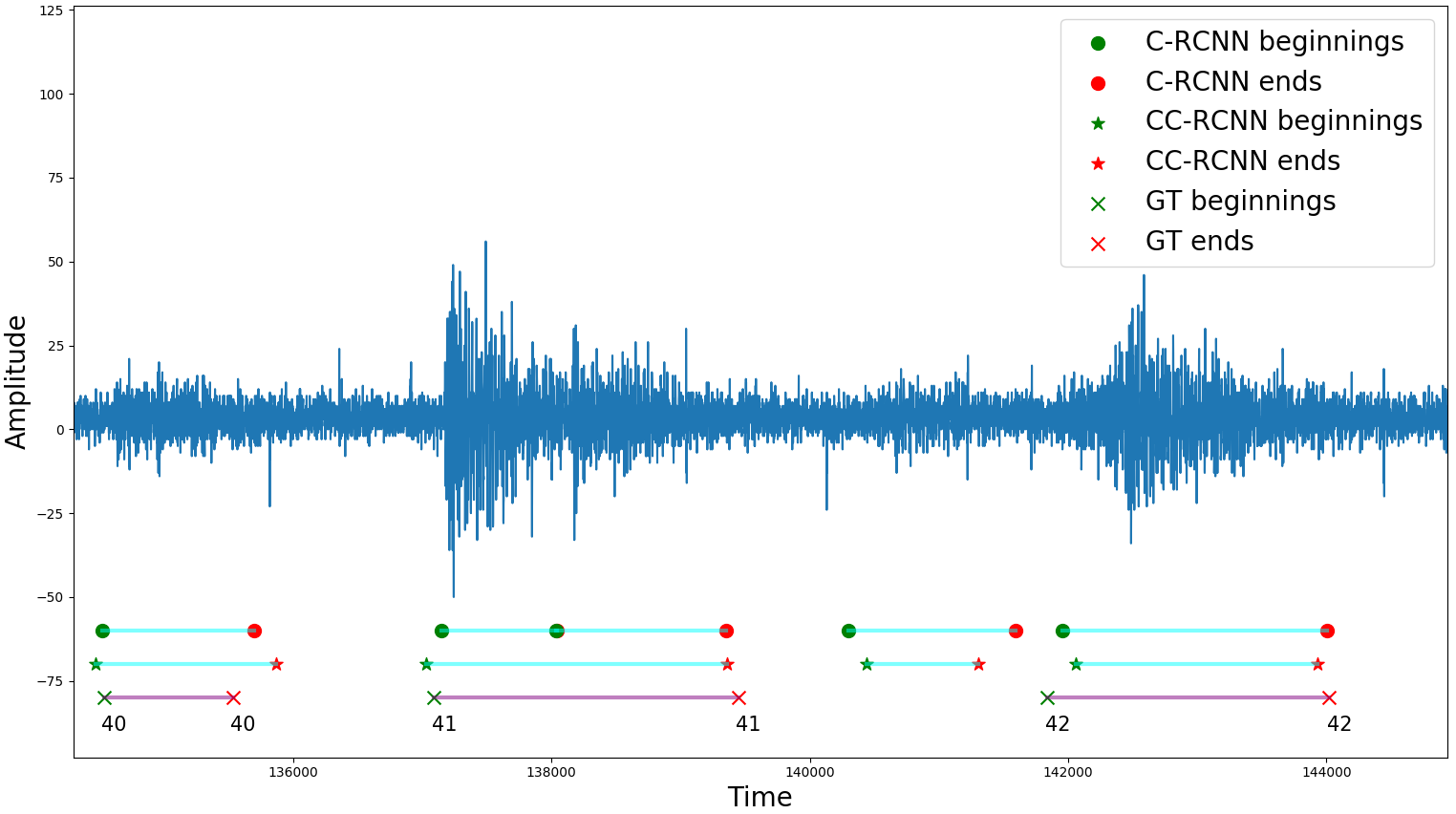}%
\label{fig:cnc_a}}
\hfil
\subfloat[]{\includegraphics[width=\linewidth]{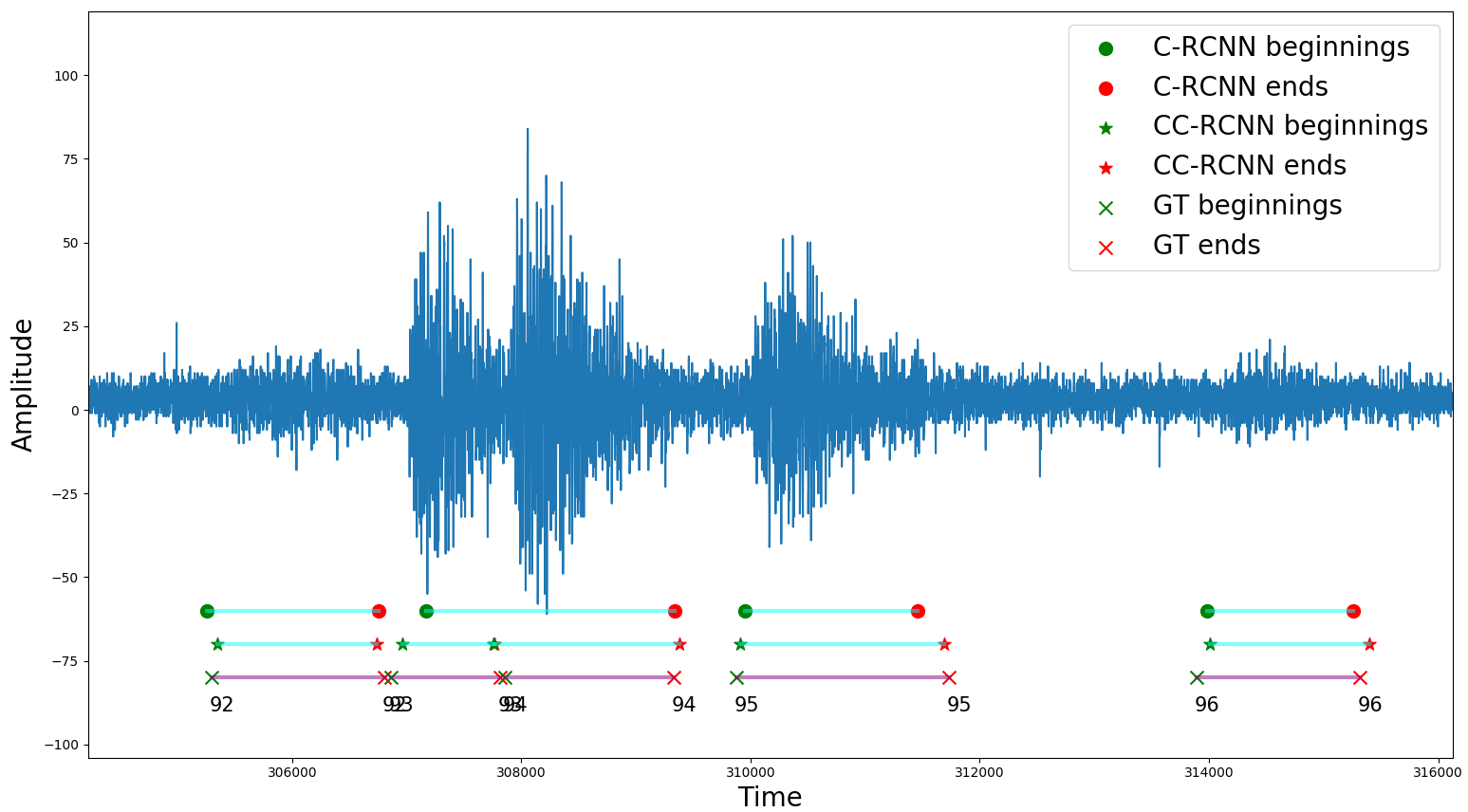}%
\label{fig:cnc_b}}
\caption{Compare the performance of C-RCNN with CC-RCNN. (a), (b) illustrate detections from two segments.
Green, red markers denote beginnings, ends, respectively. Our CC-RCNN model outperforms the C-RCNN for both examples. }
\label{fig:cnc}
\end{figure}

\subsection{Test: Robustness with Respect to Training Data Size}

Although CNN-based models have superb performances in many applications, they may require a large amount of training data to achieve low generalization
errors. Insufficient training data may lead to overfitting since the number of parameters in CNN is remarkably larger than that of other models.
Moreover, it is demanding for human annotators to amplify the training data. Thus, we conduct another sets of experiments to test the robustness of
our model with respect to different sizes of the training data.

We trained our CC-RCNN model on eight sizes of training data. We split \{0.45, 0.5, 0.55, 0.6, 0.65, 0.7, 0.75, 0.8\} of total samples as the training
set, the validation and testing sets evenly split the rest of samples. The results of robustness test are shown in Fig.~\ref{fig:trainsize}.
We observe that the accuracy benefits from the larger training set in most cases. However, even if we use 450 events out of 1000 to train our model,
we still obtain a good accuracy--AP 44.9\%. These results suggest that our DenseNet-based model yields good accuracy without enormous amount of training
data. It has a notable regularization effect achieved by the reuse of features, and they take the full advantage of skip connections to make the model robust.

\begin{table}[!b]
\centering
\begin{tabular}{ c|c|c }
& C-RCNN & CC-RCNN \\
\hline
AP@.50 & 86.2\% & 95.7\% \\
AP@.55 & 85.2\% & 94.6\% \\
AP@.60 & 81.3\% & 91.5\% \\
AP@.65 & 61.7\% & 87.5\% \\
AP@.70 & 53.8\% & 80.0\% \\
AP@.75 & 44.1\% & 72.5\% \\
AP@.80 & 35.7\% & 61.5\% \\
AP@.85 & 22.0\% & 39.2\% \\
AP@.90 & 5.5\% & 15.1\% \\
AP@.95 & 0.2\% & 0.7\% \\
\hline
AP@[.50, .95] & 47.6\% & 63.8\% \\
\end{tabular}
\caption{This table demonstrates the effect of atrous convolutions for incorporating contextual information.}
\label{table:ab_c_vs_nc}
\end{table}

\subsection{Test: Ablation Study}
We conduct ablation experiments to verify the effect of the atrous convolution blocks and the multi-scale architecture.
All ablation experiments are conducted with $\alpha = 0.5$ and $\lambda = 10$.

\subsubsection{Contextual vs Non-Contextual}

To demonstrate the importance of using atrous convolutions to incorporate contextual information, we build a non-contextual model~(C-RCNN) by
directly adding detection branches on top of $D_3-D_9$.
The improvement of performance is indicated by  AP@[.50, .95], where the contextual model outperforms the non-contextual counterpart by
13.9~points. More concrete examples are illustrate in Fig.~\ref{fig:cnc}. Event \#41 in Fig.~\ref{fig:cnc_a} is the case discussed
in Section~\ref{sec:ContextualInfo}. C-RCNN detects event \#41 as two individual events due to the second peak in the pattern.
In contrast, our contextual model, CC-RCNN, is able to capture the whole event. Moreover, the event \#93 in Fig.~\ref{fig:cnc_b} is missed by C-RCNN.
Actually, the classifier
gives positive predictions for proposals of event \#93. However, the detection of \#93 is suppressed by that of event \#94 because the
predicted beginning of event \#94 is inaccurate, which makes the IoU of these two detections above the suppression threshold.
Thus, the incorporation of contextual information for individual proposals not only reduces false detections, but also increase the localization accuracy.

\subsubsection{Multi-scale Architecture}

\begin{table}[h]
\centering
\begin{tabular}{ c|c }
& AP@[.50, .95] \\
\hline
$D_6$ & 49.5\% \\
$D_5-D_7$ & 61.0\% \\
$D_4-D_8$ & 61.2\% \\
$D_3-D_9$ & 63.8\% \\
\end{tabular}
\caption{Performances of using different detection branches. The single-scale model has the lowest accuracy. The accuracy grows as more events are covered by multi-scale anchors.}
\label{table:ab_multiscale}
\end{table}

\begin{figure}[t]
\centering
\subfloat[]{\includegraphics[width=\linewidth]{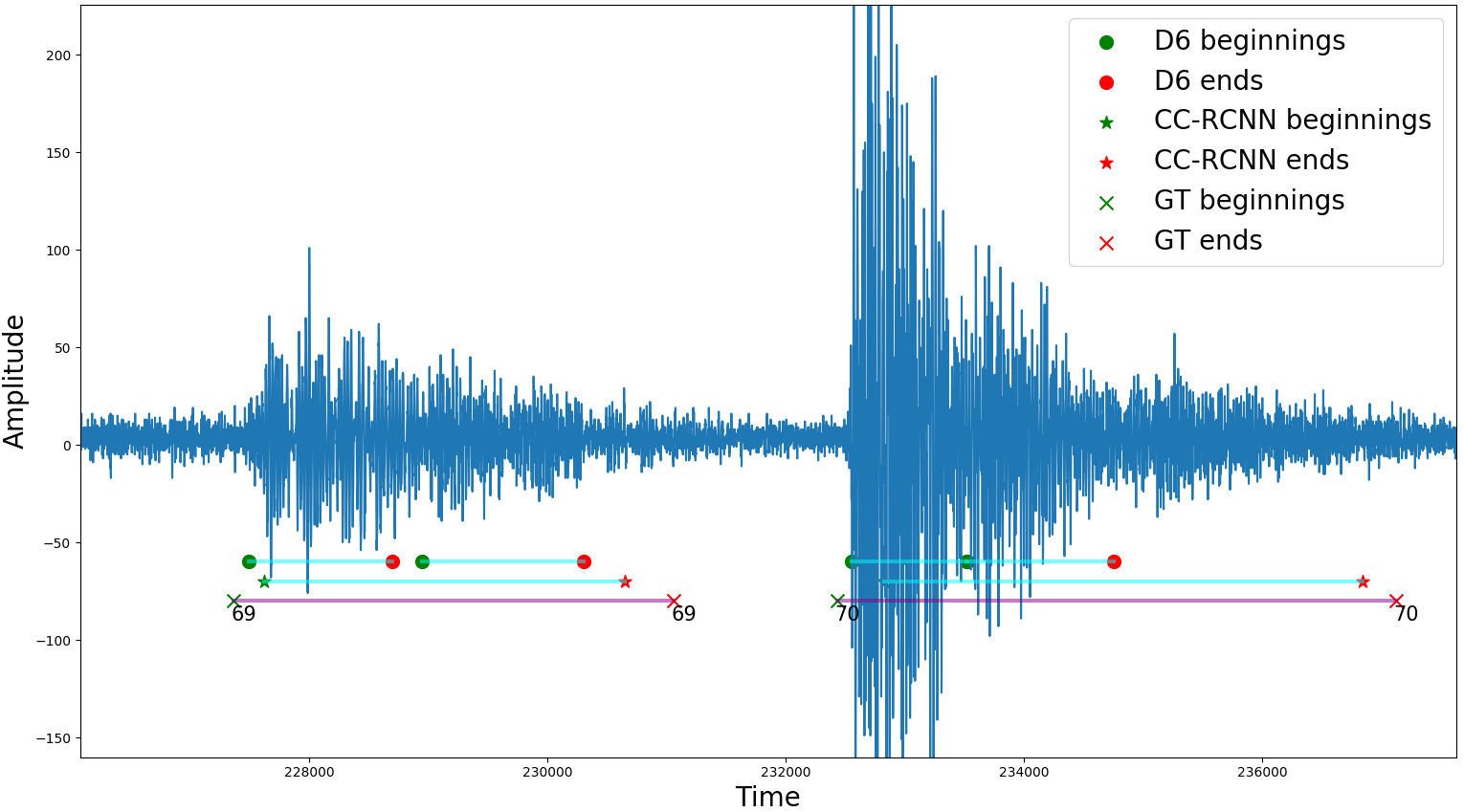}%
\label{fig:ab_multiscale_a}}
\hfil
\subfloat[]{\includegraphics[width=\linewidth]{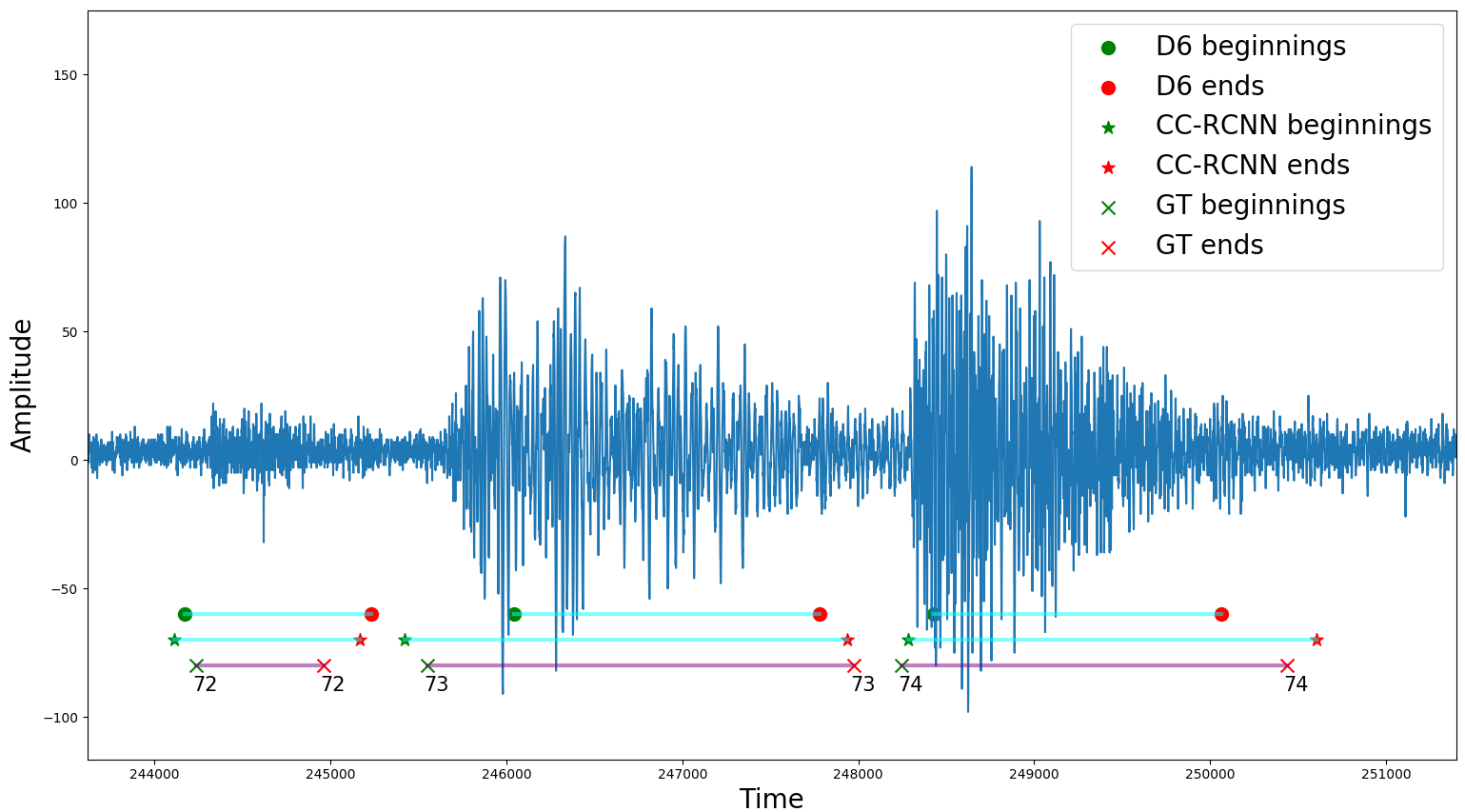}%
\label{fig:ab_multiscale_b}}
\caption{Compare detections of single-scale model using $D_6$ with our CC-RCNN model using multi-scale branches. Green, red markers denote beginnings, ends, respectively. The ground-truth is indicated at the bottom. Results show that the single-scale detector cannot capture events with lengths far away from the anchor size.}
\label{fig:ab_multiscale}
\end{figure}

\begin{figure*}[t]
\centering
\subfloat[]{\includegraphics[width=0.33\linewidth]{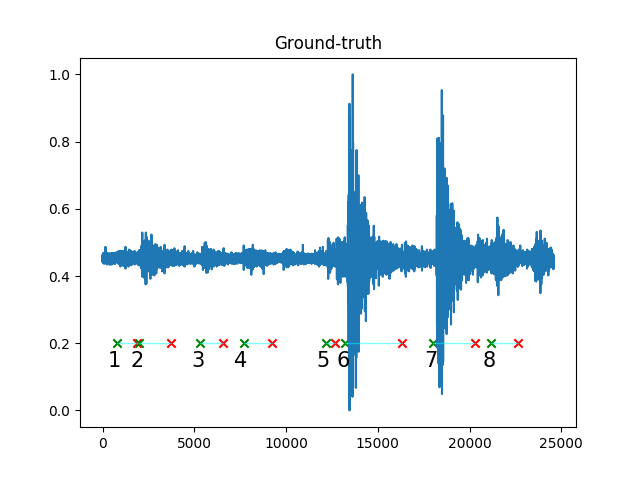}%
\label{fig:probs_ms_gt}}
\hfil
\subfloat[]{\includegraphics[width=0.33\linewidth]{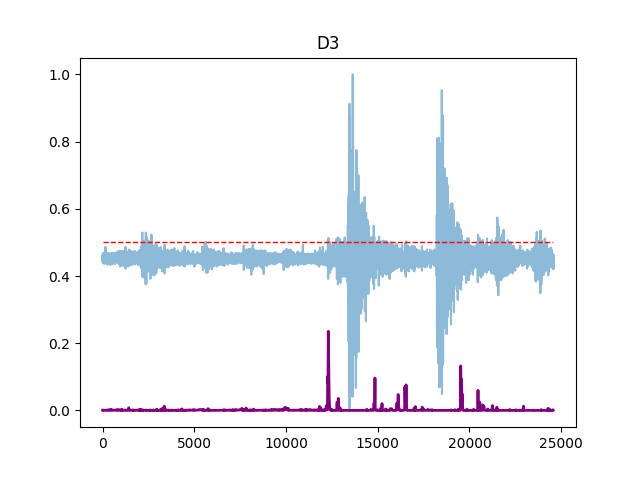}%
\label{fig:probs_ms_d3}}
\hfil
\subfloat[]{\includegraphics[width=0.33\linewidth]{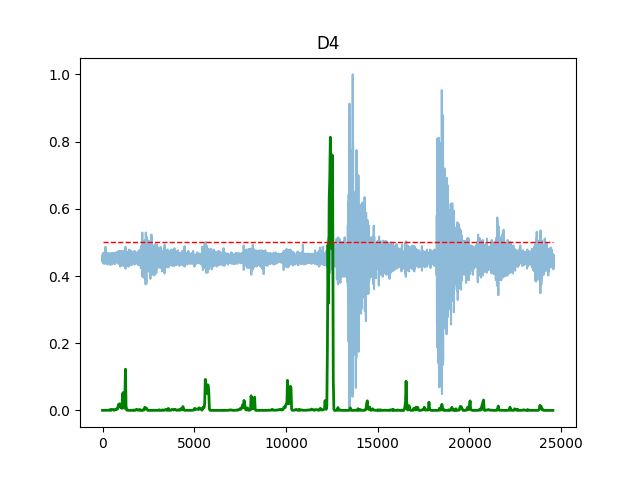}%
\label{fig:probs_ms_d4}}
\hfil
\subfloat[]{\includegraphics[width=0.33\linewidth]{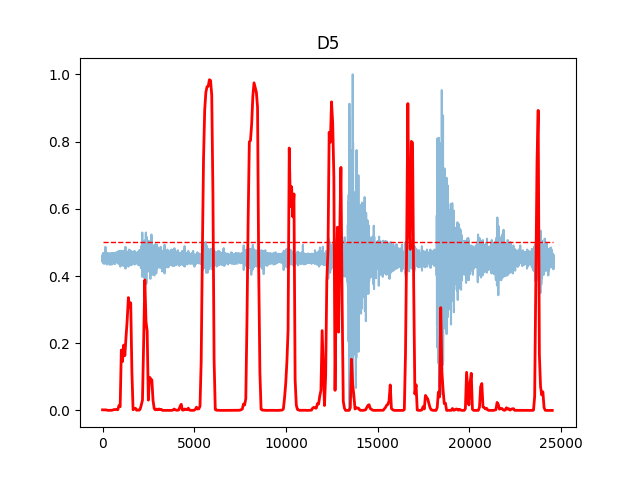}%
\label{fig:probs_ms_d5}}
\hfil
\subfloat[]{\includegraphics[width=0.33\linewidth]{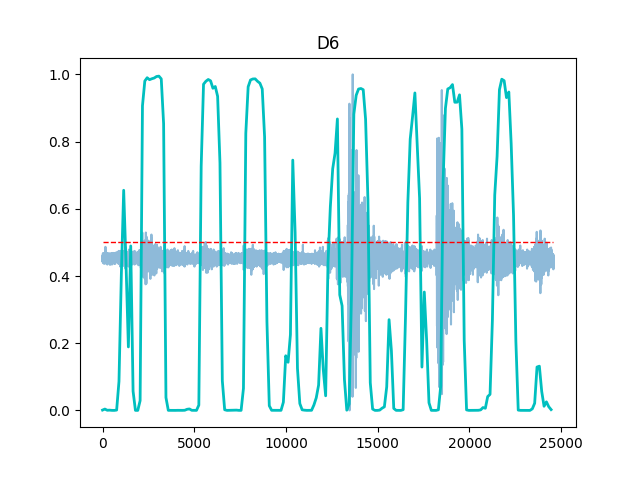}%
\label{fig:probs_ms_d6}}
\hfil
\subfloat[]{\includegraphics[width=0.33\linewidth]{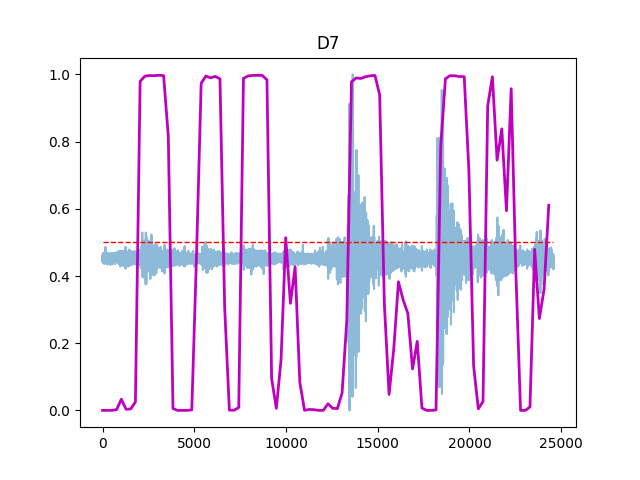}%
\label{fig:probs_ms_d7}}
\hfil
\subfloat[]{\includegraphics[width=0.33\linewidth]{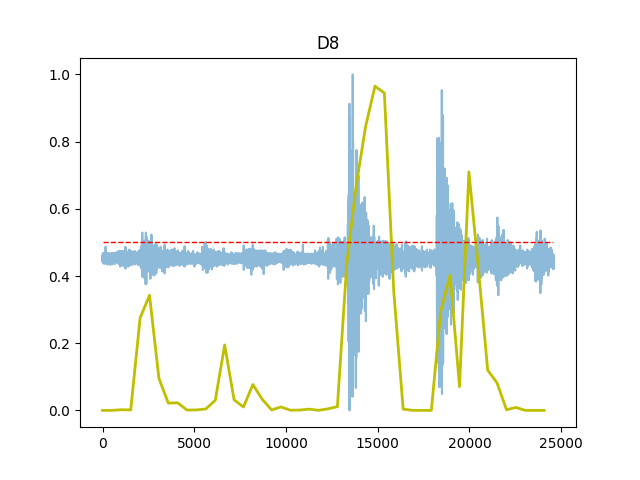}%
\label{fig:probs_ms_d8}}
\hfil
\subfloat[]{\includegraphics[width=0.33\linewidth]{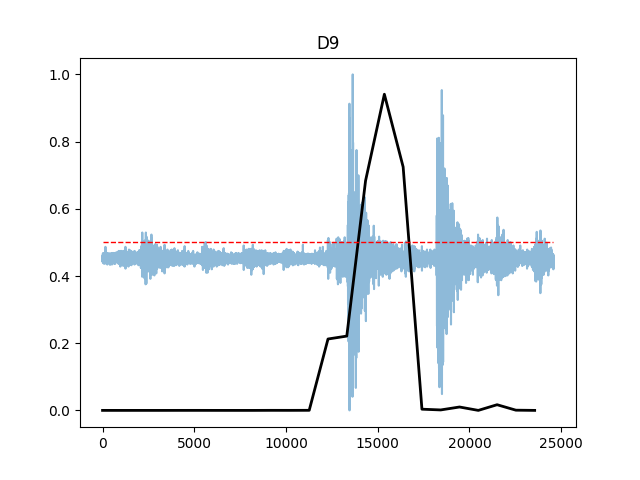}%
\label{fig:probs_ms_d9}}
\caption{Probability distributions from all detection branches. The input signals and the ground-truth are illustrated in (a).
(b)-(h) exhibit probabilities of being an event for each location of the input signal, output by $D_3-D_9$.
Signals are denoted in blue. Probabilities are denoted as solid lines and the dotted red lines indicate the detection threshold of 0.5.}
\label{fig:probs_ms}
\end{figure*}

To demonstrate the effect of the proposed multi-scale architecture, which is designed for capturing various length of events,
we compare the performance of using $D_3-D_9$ (proposed), $D_4-D_8$, $D_5-D_7$ and $D_6$ to make predictions.
The results are shown in Table~\ref{table:ab_multiscale}. The single-scale model (using $D_6$ only) achieves AP 47.3\%, and is outperformed by
the proposed multi-scale model by 14.2~points. Figure~\ref{fig:ab_multiscale} shows the detection results of two segments.
Event \#69 and \#70 in Fig.~\ref{fig:ab_multiscale_a} span about 4,000 timestamps. Event \#73 in Fig.~\ref{fig:ab_multiscale_b} spans about 3,000 timestamps. They are all poorly captured by the single-scale model since their lengths are outside the range that the anchor covers. Though, it is worthwhile to mention that the performance of the single-scale model is still significantly better than the baseline--template matching model. Since we assign positive labels to those anchors having IoU greater than 0.5 with at least one ground-truth event, $D_6$ is theoretically capable of capturing events with 512-2,048 timestamps. The majority of events in our dataset can be captured with this range.

To further demonstrate the effectiveness of each branch, we plot their probability distributions in Fig.~\ref{fig:probs_ms}.
The signals have been scaled to [0, 1]. A testing segment is shown in Fig.~\ref{fig:probs_ms_gt}, with 8 events to be captured.
The length of each event is shown in Table~\ref{table:ms_event_length}.
\begin{table}[h]
\centering
\begin{tabular}{ c|c|c|c|c|c|c|c|c }
ID & 1 & 2 & 3 & 4 & 5 & 6 & 7 & 8 \\
\hline
Length & 1102 & 1698 & 1236 & 1526 & 502 & 3058 & 2292 & 1436 \\
\end{tabular}
\caption{The length (in timestamp) of each event in Fig.~\ref{fig:probs_ms_gt}.}
\label{table:ms_event_length}
\end{table}
Figure~\ref{fig:probs_ms_d3}-\ref{fig:probs_ms_d9} display the probabilities of being an event output by the classification branch built on $D_3-D_9$.
It can be seen that none of events is captured by $D_3$ since its anchor size is only 128, and all events are beyond the scope that $D_3$ is
theoretically responsible for. Event \#5, the smallest one in this segment spanning 502 timestamps, is captured by \{$D_4, D_5, D_6$\}.
The largest event, \#6, spanning 3,058 timestamps, is captured by \{$D_6, D_7, D_8, D_9$\}. These results indicate that an event can be captured by
multiple detection branches simultaneously. However, our model is capable of suppressing redundant detections because the highest probability is
always given by the detection branch with the anchor size closest to that event. More concretely, $D_5$ gives the highest probability for
event \#5 since its anchor size is 512. $D_7$, with anchor size 2,048, gives the most confident prediction for event \#7. For event \#6, $D_7$ and $D_8$ give similar predictions since the event length is at the middle of the two anchor sizes. To sum up, Fig.~\ref{fig:probs_ms} verifies that our multi-scale architecture can effectively capture events with dramatically various lengths.

\subsection{Test: Curated and Randomly Selected Detections}
\begin{figure*}[!h]
\centering
\subfloat[]{\includegraphics[width=0.9\linewidth, height=7cm]{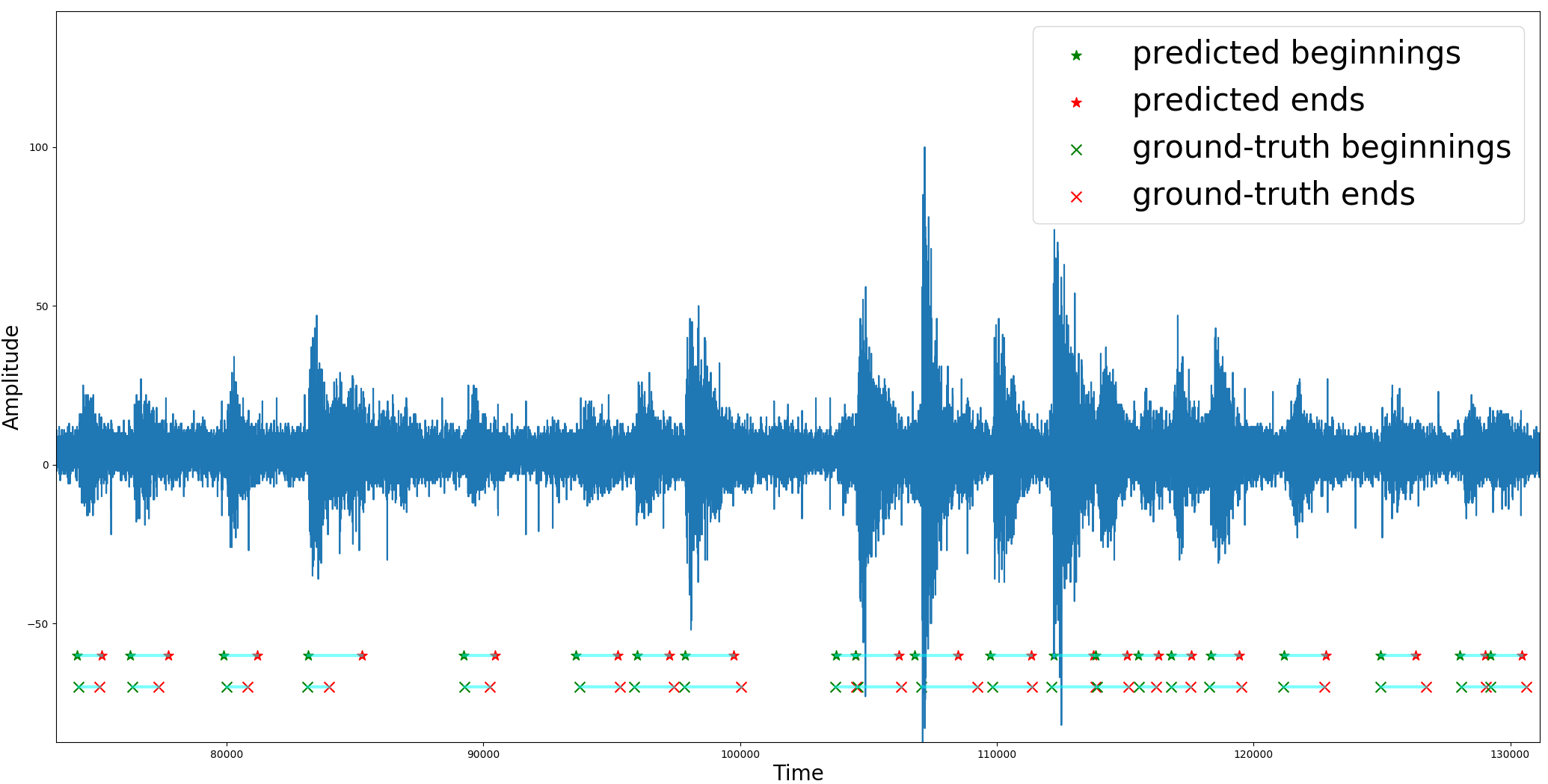}}
\hfil
\subfloat[]{\includegraphics[width=0.9\linewidth, height=7cm]{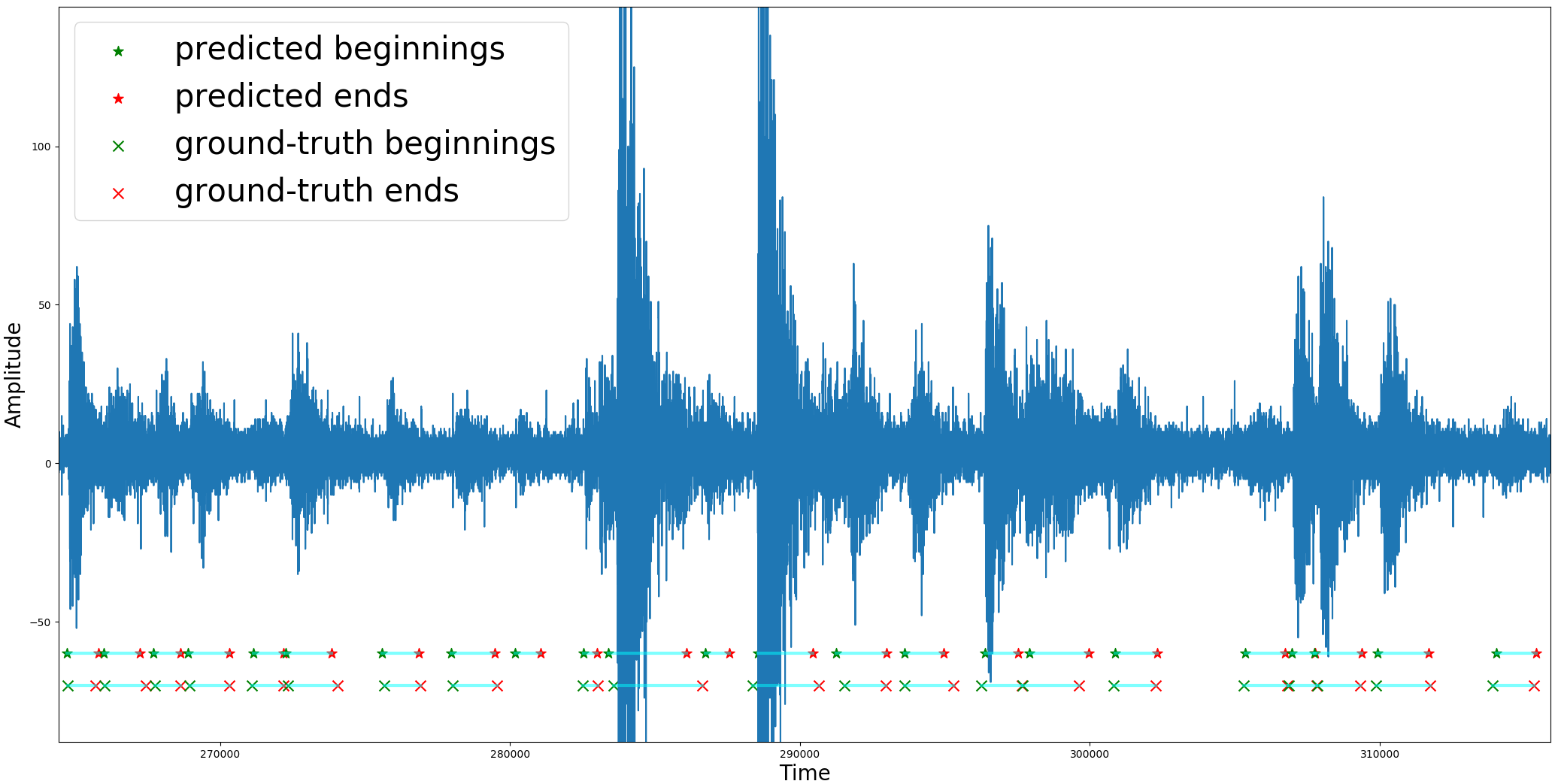}}
\caption{Curated examples. The two segments are selected because the we found the detection results are impressive.}
\label{fig:curated_1}
\end{figure*}

\begin{figure*}[!h]
\centering
\subfloat[]{\includegraphics[width=0.9\linewidth, height=7cm]{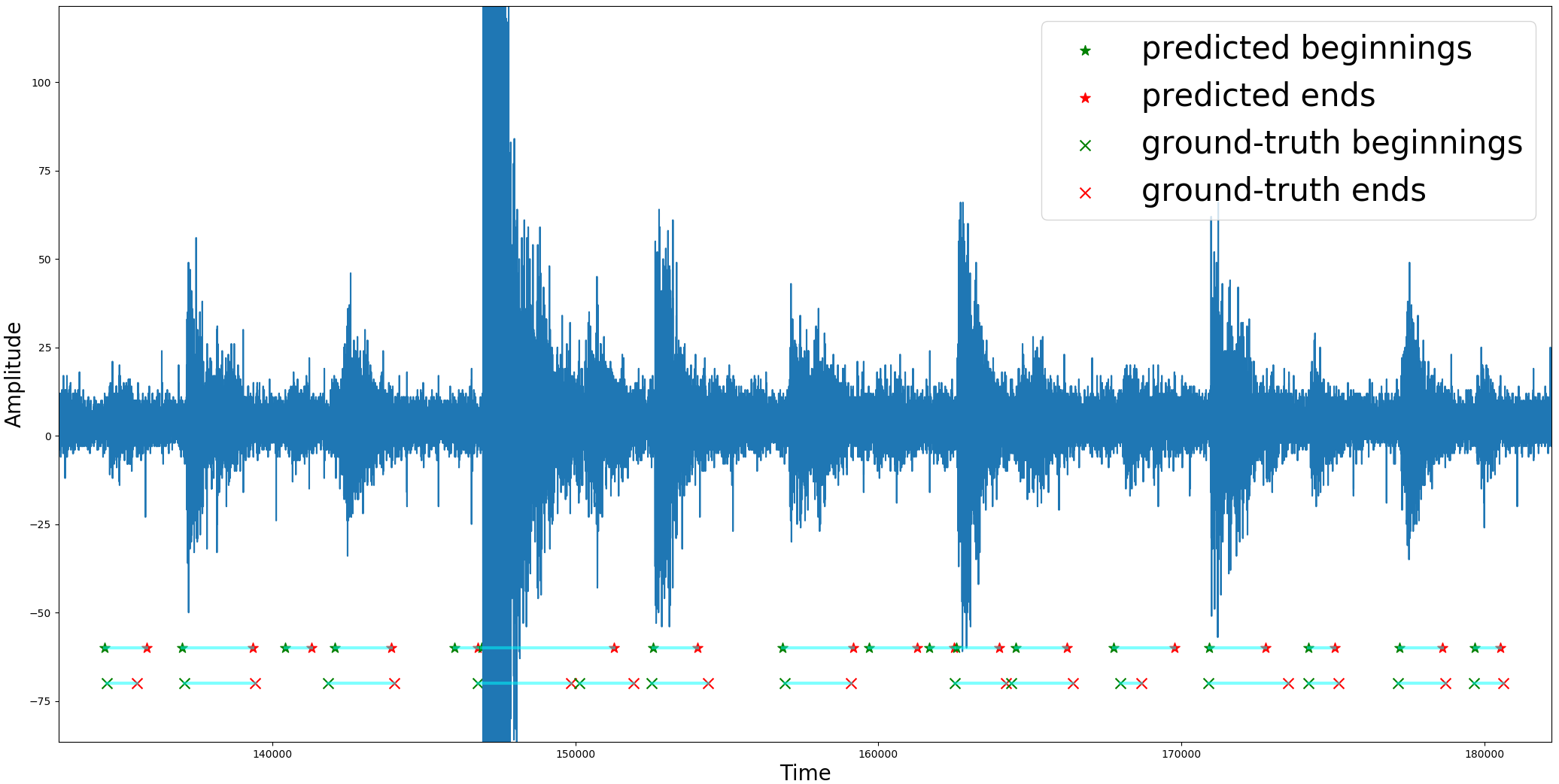}}
\hfil
\subfloat[]{\includegraphics[width=0.9\linewidth, height=7cm]{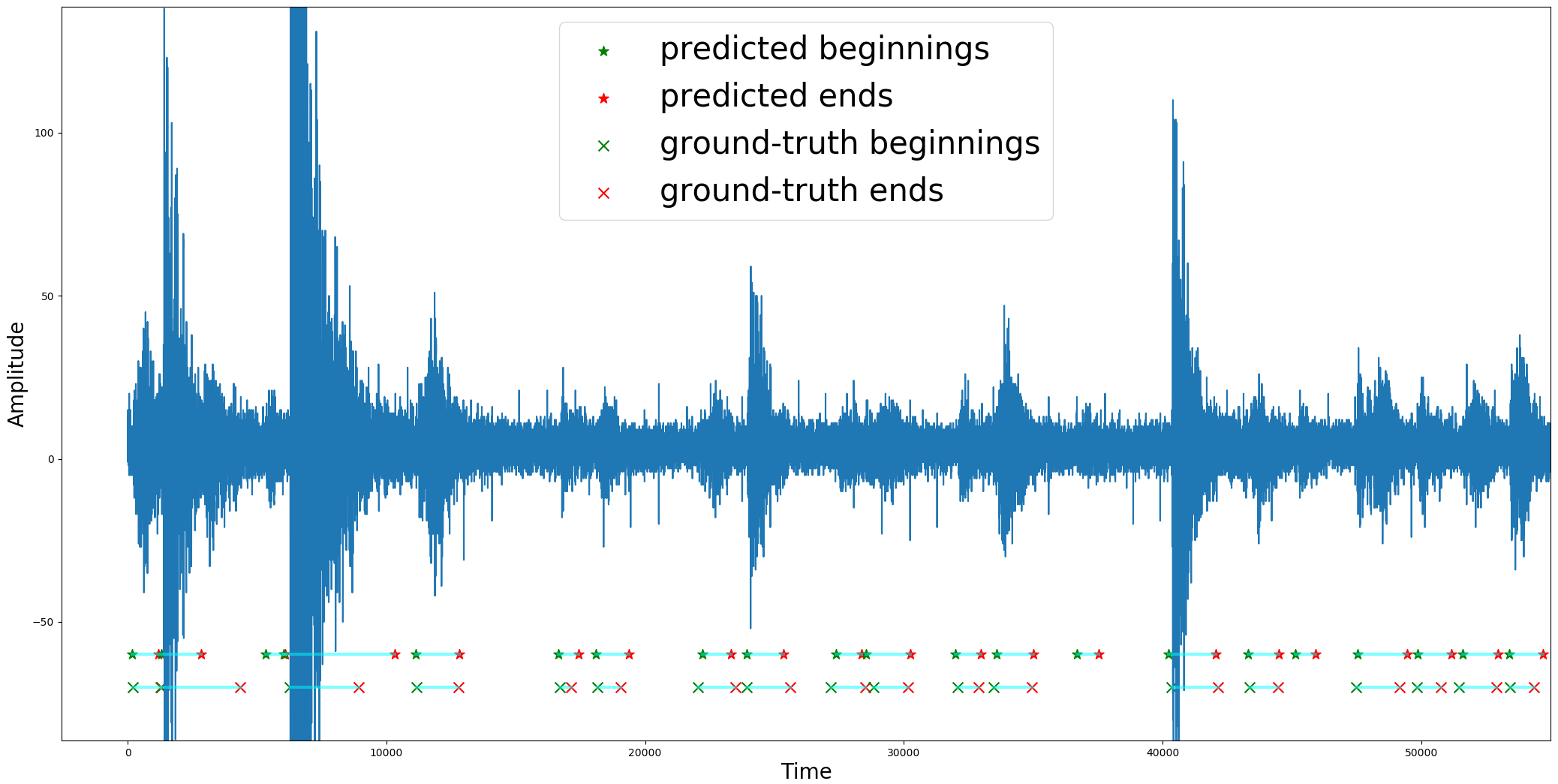}}
\hfil
\subfloat[]{\includegraphics[width=0.9\linewidth, height=7cm]{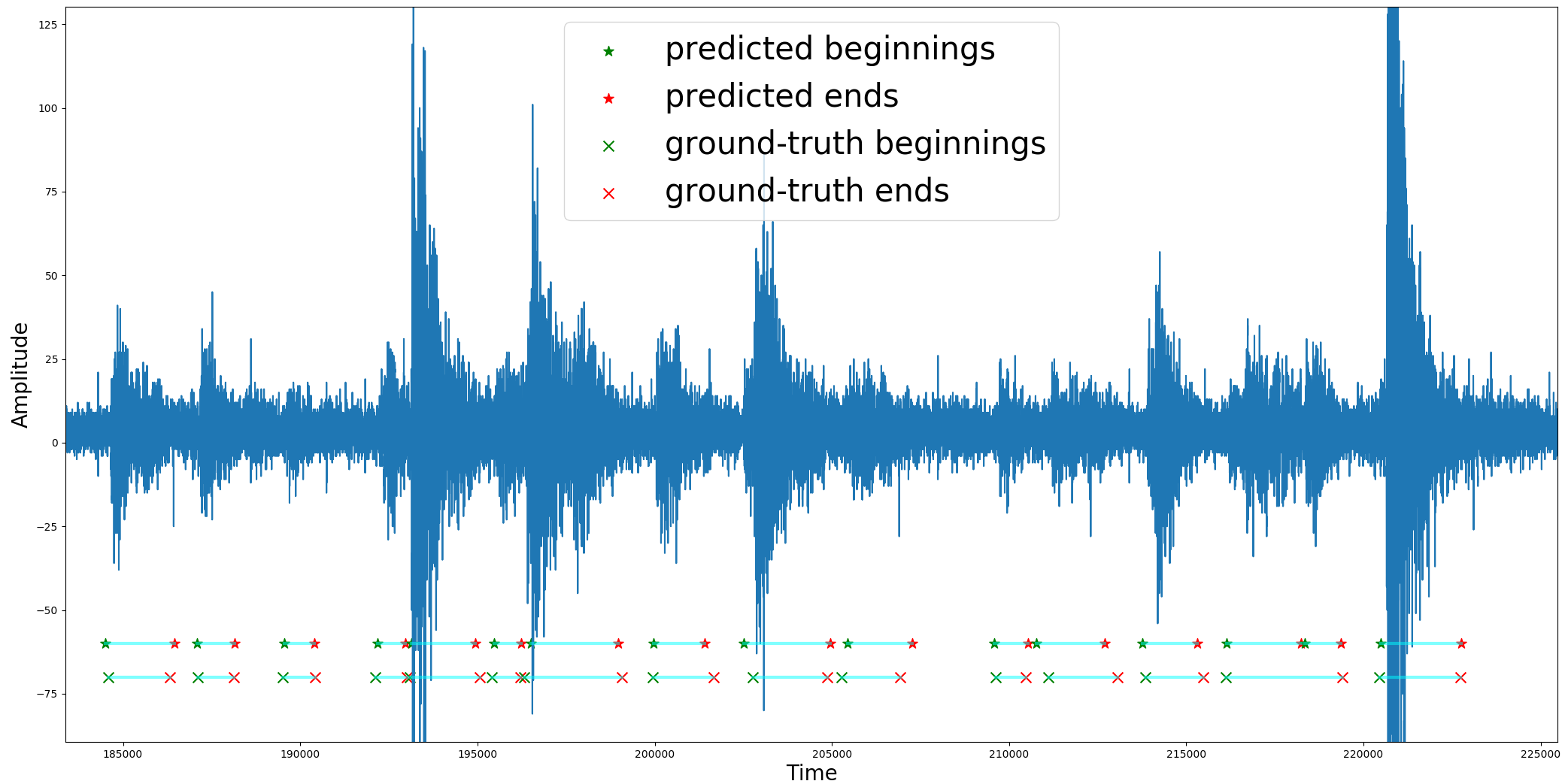}}
\caption{Randomly selected detection results.}
\label{fig:randomly_selected_1}
\end{figure*}

To further illustrate the performance of our CC-RCNN model, we present example detections of 5 long segments in Fig.~\ref{fig:curated_1} and Fig.~\ref{fig:randomly_selected_1}. All detections are plotted on the top of the ground-truth. Fig.~\ref{fig:curated_1} shows the detection results of two curated time series segments, which are selected because of the impressive detection results. The detections of three randomly selected segments are presented in Fig.~\ref{fig:randomly_selected_1}. Although they are not curated examples, we found the detection results still promising.

\section{Conclusion}
\label{sec:Conclusion}

Accurate event detection out of 1D times series seismic data is not only important but also
challenging. In this paper, we develop a novel event-wised detection method for 1D time series signals.
Specifically, a cascaded architecture is developed to generate multi-scale proposals to detect events with various lengths.
To take into account of the temporal correlation of time series data, we use atrous convolutions with different
dilation rates to enrich features of individual proposals. To help with the optimization, we share parameters for
branches built on top of multi-scale proposals. For event detection tasks in 1D time series signals, our model is state-of-the-art.
In our experimental tests, we compare our new cascaded-contextual region-based convolutional neural network model
with a standard method (template matching). Our detection accuracy is significantly higher than that obtained using template matching, especially
when the sizes of events are greatly different from one another. We also demonstrate the robustness of our new model to different sizes
of the training dataset. To conclude, our new detection model yields high performance for seismic dataset, therefore it has great potential
for event detection in various seismic applications.

\section*{Acknowledgment}
This work was co-funded by the Center for Space and Earth Science at Los Alamos National Laboratory, and the U.S. DOE Office of Fossil
Energy through its Carbon Storage Program.

\bibliographystyle{IEEEtranN}

\bibliography{bibliography}

\end{document}